\documentclass[11pt]{article}
\usepackage{makecell}
\usepackage[preprint]{acl}
\usepackage{amsmath}
\usepackage{booktabs}
\usepackage{multirow}
\usepackage{amsfonts}
\usepackage{times}
\usepackage{latexsym}
\usepackage{graphicx}
\usepackage{subcaption}
\usepackage{xcolor} 
\usepackage[most]{tcolorbox}
\usepackage[T1]{fontenc}

\usepackage[utf8]{inputenc}

\usepackage{microtype}

\usepackage{inconsolata}

\usepackage{graphicx}

\title{BalanceSFT: Improving LLM Function Calling with Balanced \\ Training Signals and Data Hardness}

\newcommand{\cmark}{$\checkmark$}  
\newcommand{\xmark}{$\times$}

\author{
 \textbf{Bingguang Hao\textsuperscript{1}$^{\ast\dagger}$},
 \textbf{ZengZhuang Xu\textsuperscript{1}$^{\ast}$},
 \textbf{Maolin Wang\textsuperscript{2}$^{\ast\dagger}$},
 \textbf{Yuntao Wen\textsuperscript{1}$^{\ast\dagger}$},
 \textbf{Yicheng Chen\textsuperscript{1}$^{\dagger}$},
 \\
 \textbf{Cunyin Peng\textsuperscript{1}},
 \textbf{Long Chen\textsuperscript{1}},
 \textbf{Dong Wang\textsuperscript{1}},
 \textbf{Xiangyu Zhao\textsuperscript{2}},
 \textbf{Jinjie Gu\textsuperscript{1}},
 \textbf{Chenyi Zhuang\textsuperscript{1}$^{\ddagger}$},
 \textbf{Ji Zhang\textsuperscript{3}$^{\ddagger}$}
\\
 \textsuperscript{1}\includegraphics[height=1em]{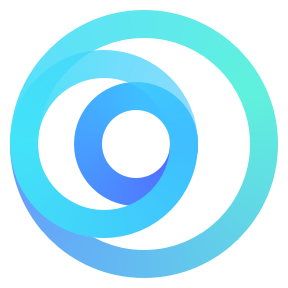}\,AWorld Team, Inclusion AI \quad
  \textsuperscript{2}City University of Hong Kong \quad
 \textsuperscript{3}Southwest Jiaotong University \\
 \texttt{\{bingguanghao7,jizhang.jim\}@gmail.com}\quad
 \texttt{\{chenyi.zcy\}@antgroup.com}
}

\begin{document}
\maketitle
{\renewcommand\thefootnote{}
\footnotetext{$^\ast$Equal contributions. $^\dagger$Work done at Ant Group.}
\footnotetext{$^\ddagger$Corresponding Author.}
\footnotetext{\ \ Project Page: \href{https://github.com/BingguangHao/BalanceSFT}{BalanceSFT}}

\begin{abstract}
While Supervised Fine-Tuning (SFT) is the prevailing method for equipping Large Language Models (LLMs) with function calling capabilities, its effectiveness is often compromised by two critical challenges: 1) \textbf{Imbalanced Training Signals}, where lengthy Chain-of-Thought (CoT) reasoning tokens dominate the training signals over concise function calls in the learning objective, and 2) \textbf{Imbalanced Data Hardness}, characterized by a scarcity of hard training examples. To overcome these limitations, we propose Balanced Supervised Fine-tuning (\textbf{BalanceSFT}), a novel framework incorporates two key components: a Self-adjusted Signal Balancing (SSB) loss that employs a learnable hyperparameter to dynamically adjust the token contributions of CoT reasoning and function calls, together with a Hard Data Re-sampling (HDR) strategy that establishes a feedback loop to selectively generate new, high-quality complex data guided by model errors. Extensive experiments demonstrate the effectiveness of our proposed BalanceSFT framework. With BalanceSFT, a 7B model achieves function calling performance on par with state-of-the-art giants like GPT-4o. Our code, models, and dataset are open-sourced.
\end{abstract}

\section{Introduction}

The rapid advancement of Large Language Models (LLMs) has marked a transformative era in natural language processing~\cite{ouyang2022training,achiam2023gpt,touvron2023llama,bai2023qwen}, demonstrating exceptional capabilities in text understanding, generation, and complex reasoning~\cite{zhong2024evaluation}. A pivotal direction in enhancing their practicality is equipping them with function calling abilities, enabling LLMs to interact with external tools and APIs to execute concrete actions and access real-world data~\cite{wang2025function,singh2024llm}. Supervised Fine-Tuning (SFT) has emerged as the standard paradigm to instill this capability, typically by training models on demonstrations that combine a Chain-of-Thought (CoT) reasoning process with the final function call~\cite{lightman2023let,guo2025deepseek,team2025kimi}.

\begin{figure}[t]
  \centering
  \includegraphics[width=\columnwidth]{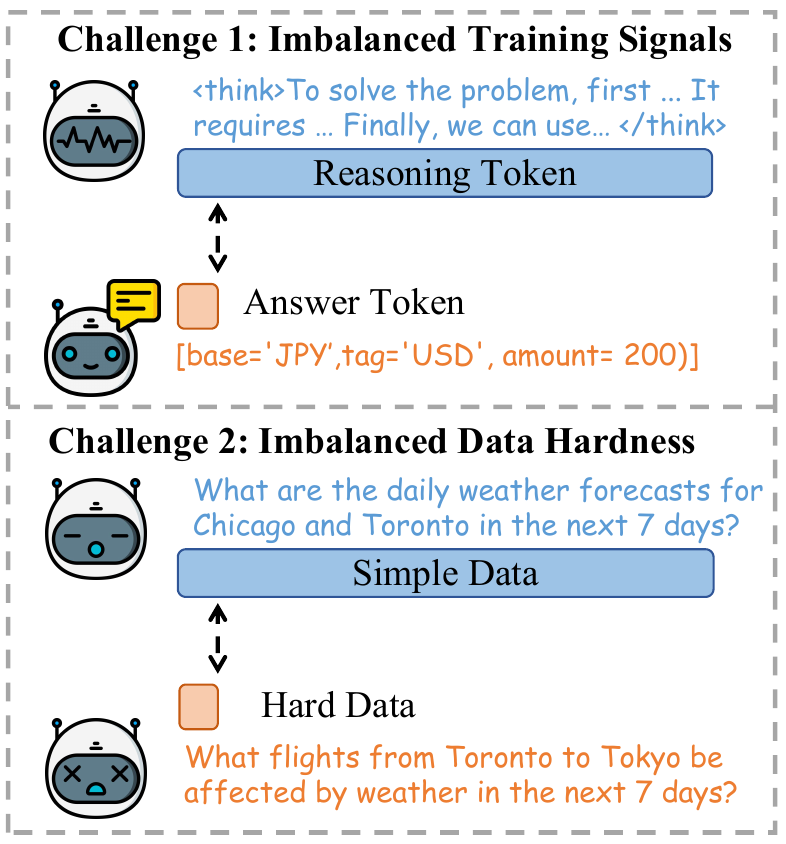}
  \caption{\textbf{Two challenges in LLM Function Calling.} \textbf{(a) Imbalanced Training Signals}: The lengthy Chain-of-Thought (CoT) tokens dominate the learning signal, overshadowing the concise but critical function call. \textbf{(b) Imbalanced Data Hardness}: The training dataset is dominated by simple examples, with a scarcity of hard cases necessary for robust performance.}
  \label{tab:intro} 
\end{figure}

Despite the remarkable progress,  conventional SFT approaches suffer from two inherent and critical imbalances that severely limit its effectiveness~\cite{ju2024mitigating}. On the one hand, as illustrated in Figure~\ref{tab:intro}, the training signals are skewed due to a token-level imbalance. In a typical function-calling sample, the lengthy CoT reasoning tokens numerically dominate the concise, yet critical, function call tokens~\cite{qian2025toolrl}. This Imbalanced Training Signal causes the model to be optimized primarily for generating plausible reasoning, often at the expense of precision in the final tool execution. On the other hand, the data distribution is characterized by an Imbalanced Data Hardness~\cite{li2023api,zhang2024xlam}. Existing training datasets are naturally dominated by simple examples, while complex and challenging cases that are crucial for robust performance are scarce. This scarcity prevents the model from effectively learning to handle edge cases and intricate scenarios~\cite{zeng2025toolace}.

To tackle the dual challenges, we propose \textbf{Balanced Supervised Fine-Tuning (BalanceSFT)}, a novel framework designed to recalibrate the SFT process. BalanceSFT incorporates two simple yet effective components: 1) A Self-adjusted Signal Balancing (\textbf{SSB}) loss function that introduces a learnable hyperparameter to dynamically balance the contribution of CoT tokens and function call tokens to the total loss, ensuring that the model does not overlook the critical execution step; 2) A Hard Data Re-sampling (\textbf{HDR}) strategy that establishes a feedback loop to generate new, high-quality complex data guided by model errors. Extensive experiments validate the effectiveness of our BalanceSFT framework. The results show that a 7B model trained with BalanceSFT achieves tool-calling performance competitive with state-of-the-art giants like GPT-4o~\cite{hurst2024gpt} on the comprehensive Berkeley Function Calling Leaderboard (BFCL)~\cite{patil2024gorilla} and other specialized benchmarks~\cite{chen2025acebench,li2023api}.

Our main contributions are threefold:
\begin{itemize}
\item We identify two critical yet overlooked limitations in standard SFT for tool learning: the imbalance in training signals between lengthy reasoning and concise function calls, and the imbalance in data hardness.
\item We propose BalanceSFT, a novel framework that introduces a Self-adjusted Signal Balancing (SSB) loss and a Hard Data Re-sampling (HDR) strategy to calibrate the learning focus.
\item We train a 7B model with the proposed BalanceSFT framework, achieving tool-calling performance comparable to the state-of-the-art model like GPT-4o.
\end{itemize}


\section{Related Work}
\label{gen_inst}

\textbf{Function Calling for LLMs.} The ability of LLMs to interact with external tools and APIs through function calling has emerged as a crucial aspect of their practical utility~\cite{wang2024executable,singh2024llm,patil2024gorilla}. This capability allows LLMs to transcend the limitations of solely processing and generating text, enabling them to ground their responses in real-world data and automate complex tasks~\cite{wang2024gta,kavathekar2025small,2025InSpire}. Early explorations in this area focused on enabling LLMs to understand descriptions of functions and generate the necessary calls with appropriate parameters~\cite{wang2023describe,zheng2023llm}. By training LLMs on datasets specifically curated for function calling scenarios, these methods aim to instill a deeper understanding of function semantics and parameter requirements~\cite{prabhakar2025apigen,zeng2025toolace}. The quality and diversity of these fine-tuning datasets play a critical role in the resulting performance of the LLMs in function calling tasks. RL has also been explored as a means to optimize LLMs for function calling~\cite{zhang2025nemotron,qian2025toolrl}. RL-based approaches often involve defining reward functions that incentivize the generation of correct and executable function calls. While RL offers the potential to directly optimize for task success, it still severe sparse reward and the lack of effective data~\cite{qu2025latent,zeng2025toolaceacer}.

\noindent\textbf{Improving Reasoning Capability via SFT.} The integration of CoT with function calling mechanisms holds substantial promise for enhancing the accuracy and reliability of tool-use~\cite{lu2024toolsandbox,kokane2025toolscan,chen2024towards}. By reasoning through the user's request and identifying the most appropriate course of action, including the selection of relevant tools and their parameters, the model can make more informed decisions about function invocation~\cite{li2025review,shen2024llm}. The explicit reasoning process provided by CoT can also improve the transparency and interpretability of the model's actions~\cite{wu2025joint,dong2025tool}. However, the integration of CoT into the training process for function calling introduces a unique challenge related to the design of the loss function~\cite{yang2025alleviating}. Traditional training methodologies, primarily SFT, typically treat the entire generated sequence (including the reasoning process and the final function call) uniformly when calculating the loss~\cite{lobo2025impact}. This can lead to an imbalance, where the potentially lengthy sequences of reasoning steps dominate the loss calculation, potentially overshadowing the importance of the final, often shorter, function call. Consequently, the model might be incentivized to generate elaborate and seemingly plausible reasoning chains~\cite{agarwal2024faithfulness,chen2025reasoning}, even if they do not ultimately lead to a correct or executable function call. This inherent tension between the verbose reasoning process and the need for a succinct and precise function call necessitates a more nuanced approach to loss function design that can effectively balance these two critical aspects of the task. Our work addresses this challenge by introducing a Self-adjusted Signal Balancing approach that explicitly considers the different roles and importance of the reasoning process and the final function execution during training.
\section{Method}
\label{headings}
In this section, we elaborate on our proposed Balanced Supervised Fine-Tuning (BalanceSFT) framework, which is designed to address the dual challenges of Imbalanced Training Signals and Imbalanced Data Hardness.

\begin{figure*}[h]
  \centering
  \includegraphics[width=\textwidth]{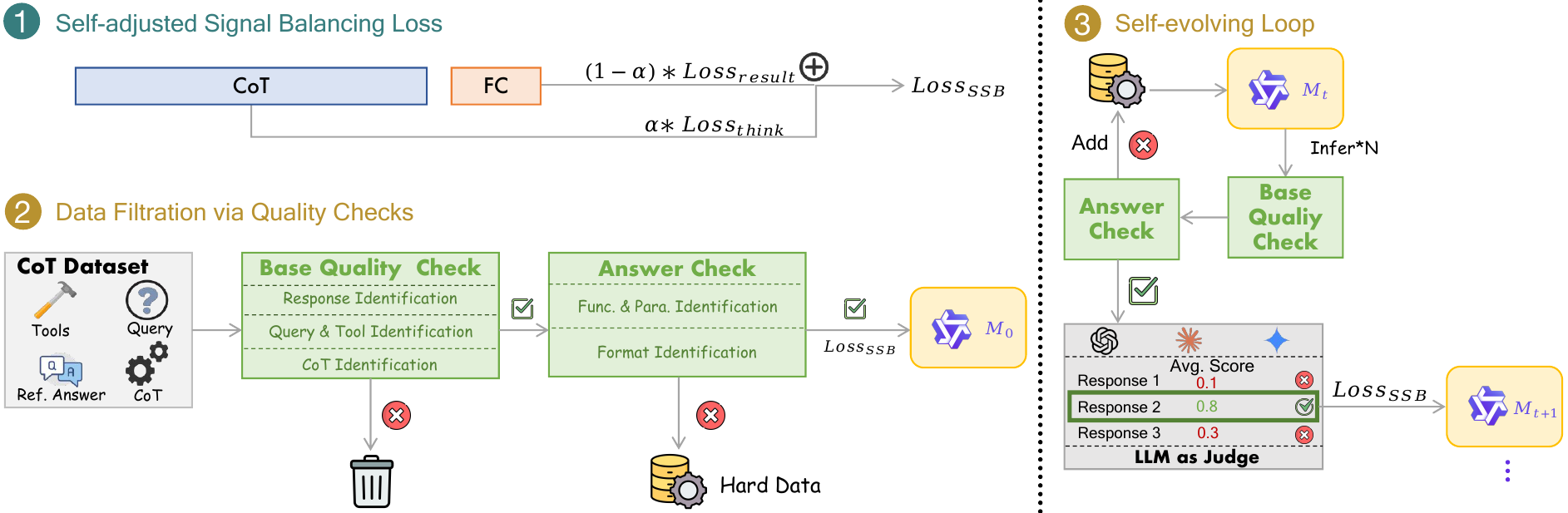}
  \caption{\textbf{Overview of the BalanceSFT framework}. It starts with a standard function call dataset, which is refined through a {Base Quality Check} and {Answer Check} to create initial training data and identify hard data. The model is first initialized via a Cold Start using the \textcolor{teal}{Self-adjusted Signal Balancing (SSB) Loss}. Subsequently, the \textcolor{brown}{Hard Data Re-sampling (HDR) strategy} creates a \textcolor{brown}{Self-evolving Loop} where the model iteratively reasons on hard cases, generates new solutions, and undergoes quality-gated retraining.}
  \label{tab:pipeline} 
\end{figure*}

\subsection{Overview}
The overall architecture of our BalanceSFT framework is illustrated in Figure~\ref{tab:pipeline}. It consists of two core components that work in synergy: 1) A Self-adjusted Signal Balancing (SSB) loss that dynamically balances the learning signals between lengthy Chain-of-Thought reasoning tokens and concise function call tokens during training, and a Hard Data Re-sampling (HDR) strategy that establishes a feedback loop to selectively generate and utilize high-quality complex data based on model errors. These two components address the signal imbalance and data hardness imbalance, respectively, forming a comprehensive solution for robust function calling capability.

\subsection{Self-adjusted Signal Balancing (SSB) Loss}


\noindent\textbf{Motivation and Analysis}. 
We conduct an intuitive analysis of a CoT Dataset. As shown in Table~\ref{tab:token}, a CoT Dataset distilled from QwQ-32B~\cite{qwq} based on the open-sourced function call dataset xlam-function-calling-60k~\cite{zhang2024xlam}, which exhibits a significant token imbalance: Chain-of-Thought reasoning segments contain approximately 10 times more tokens (mean=350.74) than function call results (mean=31.07).
This substantial disparity highlights the need to rebalance the training focus toward the short but critical function execution components.

\begin{table}[t]
\centering
\footnotesize

\begin{tabular}{lcccccc}
\toprule
\multirow{3}{*}{Dataset} & \multicolumn{2}{c}{CoT Token Len.} & \multicolumn{2}{c}{Result Token Len.} \\
\cmidrule(r){2-3} \cmidrule(r){4-5}
 & Mean & Median & Mean & Median \\
\midrule
CoT Dataset & 350.74 &248.00 &31.07 &27.00 \\
\bottomrule
\end{tabular}
\caption{Token Statistics of the CoT Dataset for function call. Note the approximately 10x higher token count for the CoT compared to the final execution.}
\vspace{-4mm} 
\label{tab:token}
\end{table}


\noindent\textbf{Formulation}. 
The output of an LLM $\mathcal{M}$ can be decomposed into two components: chain-of-thought $t$ and the final function calling execution $f$. Let $N_t$ and $N_f$ denote the token counts for these components, respectively, with $N_{\text{all}} = N_t + N_f$ representing the total tokens, and $V$ represents the size of the vocabulary. The standard SFT loss can be expressed as a weighted sum:
\begin{equation}
\begin{aligned}
L_{\text{SFT}}&=\frac{\sum_{i=1}^{N_{all}}\sum_{j=1}^{V} -p_{ij}\log p_{ij}}{N_{all}} 
\end{aligned}
\end{equation}

We decompose the loss $L_{\text{STF}}$ as following:
\begin{equation}
\begin{aligned}
L_{\text{STF}}&=\frac{N_t}{N_{\text{all}}} \cdot \frac{1} {N_t}\sum_{i=1}^{N_t}\sum_{j=1}^{V} -p_{ij}\log p_{ij}\\\
&+ \frac{N_f}{N_{\text{all}}} \cdot \frac{1}{N_f}\sum_{i=N_t+1}^{N_t+N_f}\sum_{j=1}^{V}- p_{ij}\log p_{ij}\\
&=w_t \cdot L_{\text{think}} + w_f \cdot L_{\text{result}} 
\end{aligned}
\end{equation}
where $w_t= \frac{N_t}{N_{\text{all}}}, w_f=\frac{N_f}{N_{\text{all}}}$, thus $w_t + w_f=1$. 
Let $L_{\text{think}}$ and $L_{\text{result}}$ represent the loss for reasoning and function call components, respectively.

The statistical analysis from Table~\ref{tab:token} reveals that $N_t \gg N_f$, which implies $w_t \gg w_f$ during standard training. This naturally leads to the model being optimized primarily for generating plausible reasoning at the expense of precise function execution. To address this inherent imbalance, we propose the Self-adjusted Signal Balancing (SSB) loss:
\begin{equation}
     L_{\text{SSB}} = \alpha\cdot L_{\text{think}} + (1-\alpha) \cdot L_{\text{result}}
\end{equation}
where $\alpha$ is a learnable parameter with $\alpha \in [0, 1]$. The introduction of $\alpha$ allows the model to dynamically adjust its focus between reasoning depth and execution accuracy based on performance requirements. In practice, we set $\alpha$ as a trainable parameter to avoid extensive hyperparameter tuning.

\subsection{Hard Data Re-sampling (HDR) Strategy}

The core objective of the HDR strategy is to address the {Imbalanced Data Hardness} issue through a carefully designed self-evolving data refinement loop. This process can be formally described as a sequence of iterative data filtration and augmentation operations. Let \(\mathcal{D}_{\text{initial}}\) denote the initial training dataset. Each data point \(d_i = (q_i, c_i, f_i)\) consists of a user query \(q_i\), a CoT reasoning sequence \(c_i\), and a function call \(f_i\).

\noindent\textbf{Data Filtration via Quality Checks}. 
The dataset undergoes a two-stage filtration process defined by boolean functions \(\text{BQC}(\cdot)\) and \(\text{AC}(\cdot)\) for Base Quality Check (including response, query, and CoT identification) and Answer Check (including function, parameter, and format verification), respectively:
\begin{equation}
\begin{aligned}
\mathcal{D}_{\text{qualified}} &= 
\{\, d_i \in \mathcal{D}_{\text{initial}} \mid \text{BQC}(d_i)\!\land\!\text{AC}(d_i) \,\},\\[2pt]
\mathcal{D}_{\text{hard}}      &= 
\{\, d_i \in \mathcal{D}_{\text{initial}} \mid \text{BQC}(d_i)\!\land\!\neg\text{AC}(d_i) \,\}.
\end{aligned}
\end{equation}

Here, \(\mathcal{D}_{\text{qualified}}\) is used for the {Cold Start} phase to initialize the model \(\mathcal{M}_0\), while \(\mathcal{D}_{\text{hard}}\) is reserved for iterative self-evolution.

\noindent\textbf{Self-Evolving Loop}. The model at iteration \(t\), denoted as \(\mathcal{M}_t\), is used to reason on the current hard dataset \(\mathcal{D}_{\text{hard}}^t\), generating a new set of candidate responses \(\mathcal{R}^{t+1}\):
\begin{equation}
\mathcal{R}^{t+1} = \{ \mathcal{M}_t(q_i) \mid \forall q_i \in \mathcal{D}_{\text{hard}}^t \}
\end{equation}
Each candidate response \(r_j \in \mathcal{R}^{t+1}\) is evaluated by an ensemble of LLM judges \(\mathcal{J} = \{ J_1, J_2, \dots, J_k \}\) (e.g., Gemini-2.5-Pro~\cite{comanici2025gemini}, GPT-4o~\cite{hurst2024gpt}, Claude-3.5 sonnet~\cite{TheC3}) via majority voting. The scoring function \(\text{Score}(r_j)\) aggregates their judgments:
\begin{equation}
\text{Score}(r_j) = \frac{1}{k} \sum_{m=1}^{k} \mathbb{I}\left(J_m(r_j) \ge \tau\right)
\end{equation}
where \(\mathbb{I}(\cdot)\) is the indicator function and \(\tau\) is a passing threshold. Responses meeting the quality standard form the new high-quality dataset:
\begin{equation}
\mathcal{D}_{\text{new}}^t = \{ r_j \in \mathcal{R}^{t+1} \mid \text{Score}(r_j) = 1 \}
\end{equation}

The model is then fine-tuned on this newly generated data:
\begin{equation}
\mathcal{M}_{t+1} = \text{SFT}\left(\mathcal{M}_t, \mathcal{D}_{\text{new}}^t\right)
\end{equation}

Simultaneously, the hard dataset is updated by retaining failed samples:
\begin{equation}
\mathcal{D}_{\text{hard}}^{t+1} = \{ d_i \in \mathcal{D}_{\text{hard}}^t \mid \text{Score}(\mathcal{M}_t(q_i)) < 1 \}
\end{equation}

This iterative process continues until convergence criteria are met: $\mathcal{D}_{\text{hard}}^{t+1} = \emptyset$ {or} $t = T_{\text{max}}$.

The entire HDR strategy establishes a virtuous cycle where model improvement and data quality enhancement reinforce each other, effectively resolving the data hardness imbalance through the iterative refinement process:
\begin{equation}
\mathcal{M}_{t+1}, \mathcal{D}_{\text{hard}}^{t+1} = \text{HDR-Iteration}(\mathcal{M}_t, \mathcal{D}_{\text{hard}}^t)
\end{equation}

\section{Experiments}

The experiments of this paper are divided into five parts. The first part introduces the Experiment setting. The second part elaborates on the Data Preparation. In the third part, we compare the BalanceSFT models with current powerful models and function call specialized models on the BFCL benchmark ~\cite{patil2024gorilla} and other specialized benchmarks. In the fourth part, we present an ablation study, while the final part provides further insights into the proposed method.

\subsection{Experiment Setting}

\noindent\textbf{Dataset.} As shown in Table~\ref{statisticsdatasets}, our training dataset comprises two components: xlam-function-calling-60k~\cite{zhang2024xlam} and Open-Agentic-tool-use~\cite{Open-Agentic-tool-use}. xlam-function-calling-60k contains 60000 samples from 3,673 APIs across 21 categories, while Open-Agentic-tool-use includes 30000 samples from 129 APIs across 8 categories. The former focuses on single-turn tool-use to strengthen the model's basic capabilities, while the latter focuses on multi-turn tool-use to address complex scenario requirements. The training dataset we adopted covers thousands of APIs from multiple categories, helping the model learn cross-domain API matching logic and improve its generalization ability. The distribution of the diverse and comprehensive API resources is shown in Figure~\ref{fig:APIategories}, and the details of the dataset are further illustrated and analyzed in Appendix~\ref{Dataset}. 

\noindent\textbf{Benchmarks.} 
To assess our methodology, we choose a highly representative benchmark, the Berkeley Function Calling Leaderboard (BFCL) ~\cite{patil2024gorilla}. For BFCL, we conduct evaluations on BFCLv3, which contains single-turn and multi-turn function calling scenarios. Performance across all subsets of BFCL is reported in terms of accuracy (\%). We further evaluate our model on two specialized tool-use benchmarks, ACEBench~\cite{chen2025acebench} and APIBank~\cite{li2023api}.


\noindent\textbf{Backbones.} 
To demonstrate the effectiveness and general applicability of our BalanceSFT framework, we select Qwen2.5-Coder-7B-Inst~\cite{hui2024qwen2} as our primary base model. To demonstrate the effectiveness of our method, we also conduct additional experiments using Llama-3.2-3B-Inst~\cite{touvron2023llama}.

\noindent\textbf{Baselines.} 
For comprehensive comparison, we include both open-source and closed-source models: open-source models include Moonshotai-Kimi-K2-Instruct~\cite{team2025kimi}, DeepSeek-R1-0528~\cite{guo2025deepseek}, Qwen3-235B-A22B-Instruct-2507~\cite{yang2025qwen3}, ToolACE-MT (8B)~\cite{zeng2025toolace}, and BitAgent-8B~\cite{Bitagent}; closed-source models include GPT-5~\cite{openai2025gpt}, GPT-4o~\cite{hurst2024gpt}, Gemini-2.5-Pro~\cite{comanici2025gemini}, and Grok-4~\cite{grok4}.

\noindent\textbf{Implementation Details.} 
Our implementations are based on the training framework LLaMA Factory~\cite{zheng2024llamafactory}. To ensure fair comparison, we maintain consistent hyperparameters across all experiments: batch size of 512, learning rate of $4 \times 10^{-5}$, warm-up rate of 0.05. The SSB loss hyperparameter $\alpha$ is initialized to 0.7 and set as a learnable parameter. During inference, we set the number of inferences to 3 with a temperature 0.7. In HDR, for the LLM-as-judge phase, we use Gemini-2.5-Pro~\cite{comanici2025gemini}, GPT-4o~\cite{hurst2024gpt},
Claude-3.5 Sonnet~\cite{TheC3} as judges with the same temperature setting at 0.7, threshold $\tau=0.5$, and scoring range [0, 1]. The maximum iteration $T_{max}$ is 10. All trainings run on 8 NVIDIA H20 GPUs.


\begin{table*}[h]
\small
\renewcommand{\arraystretch}{1.0}
\centering
\begin{tabular}{@{}lccccccccccc@{}}
\toprule
&& \multicolumn{5}{c}{\textbf{Multi-Turn}}& \multicolumn{3}{c}{\textbf{Single-Turn}}\\
\cmidrule(lr){3-7}\cmidrule(lr){8-10}
\textbf{Models}&\begin{tabular}[c]{@{}c@{}}\textit{Parameter}\\ \textit{Counts}\end{tabular}
&\multicolumn{1}{c}{\textit{Overall}} &
  \multicolumn{1}{c}{\textit{Base}} &
  \multicolumn{1}{c}{\begin{tabular}[c]{@{}c@{}}\textit{Miss}\\ \textit{Func}\end{tabular}} &
  \multicolumn{1}{c}{\begin{tabular}[c]{@{}c@{}}\textit{Miss}\\ \textit{Param}\end{tabular}} &
  \multicolumn{1}{c}{\begin{tabular}[c]{@{}c@{}}\textit{Long}\\ \textit{Context}\end{tabular}}&
  \multicolumn{1}{c}{\textit{Overall}} &
  \multicolumn{1}{c}{\textit{Non-Live}} &
  \multicolumn{1}{c}{\textit{Live}} \\ 
  \midrule
\multicolumn{10}{c}{Closed-Sourced Models} \\
\midrule
\textbf{GPT-5-2025-08-07}&/&28.50	&33.50	&29.50	&23.00	&28.00	&65.59&72.92	&58.25\\
\textbf{GPT-4o-2024-11-20}&/&42.50	&\underline{55.50}	&34.50	&29.00	&\textbf{51.00}&77.21	&83.88	&70.54\\
\textbf{Gemini-2.5-Pro}&/&25.00	&25.50	&26.00	&24.50	&24.00&74.50	&85.04	&63.95\\
\textbf{Gemini-2.5-Flash}&/ &33.62	&36.00	&32.50	&32.50	&33.50&77.78	&83.10	&72.46\\
\textbf{o3-2025-04-16}&/&38.38&44.00	&40.50	&31.50	&37.50&53.01	&39.98	&66.03	\\
\textbf{Amazon-Nova-Pro-v1:0}&/&34.75	&42.50	&24.50	&27.50	&44.50	&81.78    &85.25	&78.31\\
\textbf{Grok-4-0709}&/&36.12	&44.00	&31.00	&26.00	&43.50	&79.80&85.21	&74.39\\
  \midrule
\multicolumn{10}{c}{Open-Sourced Models} \\
\midrule
\textbf{Moonshotai-Kimi-K2-Inst}&1000B&41.25	&51.00	&\underline{43.00}	&31.00	&40.00	&80.80&84.02	&77.57\\
\textbf{DeepSeek-R1-0528}&671B &\underline{44.50}	&54.50	&41.00	&\underline{36.50}	&46.00&78.22&75.73	&\textbf{80.90}\\	
\textbf{Qwen3-235B-A22B-Inst-2507}&235B&39.62	&53.50	&34.50	&27.50	&43.00	&\underline{83.37}&\textbf{90.12}	&76.61	\\
\textbf{Llama-4-Maverick}&400B&17.88	&23.50	&18.00	&14.00	&16.00 &80.90 &88.15	&73.65\\
\textbf{Qwen3-8B}&8B&24.12	&26.00	&29.00	&20.50	&21.00&83.20	&\underline{88.60}	&77.79\\	
\textbf{BitAgent-8B}&8B&37.75	&46.50	&37.50	&24.00	&43.00&81.71	&87.33	&76.09\\
\textbf{watt-tool-8B}&8B&37.88	&45.50	&39.00	&24.00	&43.00&81.71	&87.54	&75.87\\
\textbf{ToolACE-MT}&8B & 40.25& \textbf{57.50} & 31.50 &34.00 & 38.00 &78.23& 84.94& 71.52  \\
\textbf{Qwen2.5-Coder-7B-Inst}&7B &3.88&6.00&3.50&2.00&4.00&76.82&83.88&69.75\\
\midrule
\multicolumn{10}{c}{Ours} \\
\midrule
\textbf{BalanceSFT-7B}&7B&\textbf{47.00} &53.50& \textbf{47.00}& \textbf{41.00} &\underline{46.50}&\textbf{84.00}&88.29&\underline{79.70}\\
\bottomrule
\end{tabular}
\vspace{-2mm}
\caption{\label{tab:bfcl-overall} Performance on BFCL (last updated August 26, 2025), with all metrics calculated using the official script. The best result within each category is highlighted in \textbf{bold}.The second-best results are \underline{underlined}.}
\end{table*}

\begin{table}[ht]
\small
\centering
\begin{tabular}{lcccc}
\toprule
Models & Non-Live & Live  & Overall \\
\midrule
Qwen2.5-Coder-7B-Inst & 83.88 & 69.75 &76.82  \\
w/ Strategy CoT  & 87.00 & 66.81  & 76.91 \\
w/ Natural CoT & 85.67 & 71.73  & 78.70\\
\bottomrule
\end{tabular}
\caption{Performance of the prescribed strategy mode and the natural reasoning mode on single-turn subset of BFCL. The two models are trained on Qwen2.5-Coder-7B-Inst with the same training hyperparameters.}
\label{tab:data_Coder}
\end{table}

\subsection{Data Preparation}\label{data_prep}



We construct two CoT datasets: (1) Strategy CoT, generated by GPT-4o following prescribed reasoning strategies, and (2) Natural CoT, generated by QwQ-32B~\cite{qwq} based on its natural reasoning process (implemented with vLLM~\cite{kwon2023efficient}, temperature = 0.1).
We conduct standard SFT on Qwen2.5-Coder-7B-Inst using each dataset. As shown in Table~\ref{tab:data_Coder}, we observe that Natural CoT consistently outperforms Strategy CoT. Therefore, we adopt Natural CoT for cold-start data generation and employ self-inference in the Hard Data Re-sampling (HDR) stage to maintain model-consistent reasoning patterns.

\begin{table}[h]
\centering
\small
\begin{tabular}{lccc}
\toprule
Dataset                   & Samples & Apis &Categories \\
\midrule
xlam-function-calling-60k & 60000   & 3673 & 21                             \\
Open-Agentic-tool-use &   30000      & 129     &    8 \\
\bottomrule
\end{tabular}
\caption{Statistics of the xlam-function-calling-60k and Open-Agentic-tool-use dataset.}
\vspace{-4mm} 
\label{statisticsdatasets}
\end{table}

\subsection{Main Results}



The performance of BalanceSFT-7B (built upon Qwen2.5-Coder-7B-Inst) on the BFCL benchmark is summarized in Table~\ref{tab:bfcl-overall}. We report accuracy across the main subsets using the official BFCL evaluation script.

\noindent\textbf{Multi-Turn Evaluation.} The Multi-Turn subset evaluates a model’s ability to handle complex, continuous conversations. As shown in Table~\ref{tab:bfcl-overall}, BalanceSFT-7B demonstrates the strongest overall capability, leading the Overall score with 47.00. It also ensures optimal performance in the key failure mode categories: Miss Function at 47.00 and Miss Parameter at 41.00, indicating the ability to keenly detect the absence of functions and required parameters in the dialogue flow. In contrast, the Long Context metric, which tests memory and coherence over conversation history, is led by the closed-source model GPT-4o-2024-11-20~\cite{hurst2024gpt} at 51.00, with BalanceSFT-7B being the second best at 46.50. 

\noindent\textbf{Single-Turn Evaluation.} The Single-Turn subset focuses on isolated prompts under Non-Live and Live settings. As reported in Table~\ref{tab:bfcl-overall}, BalanceSFT-7B  maintains its competitive edge by achieving the highest Overall score of 84.00. The Non-Live metric is led by Qwen3-235B-A22B-Instruct-2507~\cite{yang2025qwen3} with the highest score of 90.12. For the Live metric, simulating the real-world interactions, DeepSeek-R1-0528\cite{guo2025deepseek} performs the best at 80.90, with BalanceSFT-7B coming in a close second at 79.70.

Despite being a small model, BalanceSFT-7B outperforms larger proprietary and specialized models, revealing that our BalanceSFT strategy achieves superior performance, particularly in the multi-turn function calling scenario. These findings demonstrate that BalanceSFT-7B can accurately analyze system status and remains highly sensitive to key functions and parameters within context.


\begin{figure*}[!t]
    \centering
    \begin{subfigure}[b]{0.49\textwidth}
        \centering
        \includegraphics[width=\textwidth]{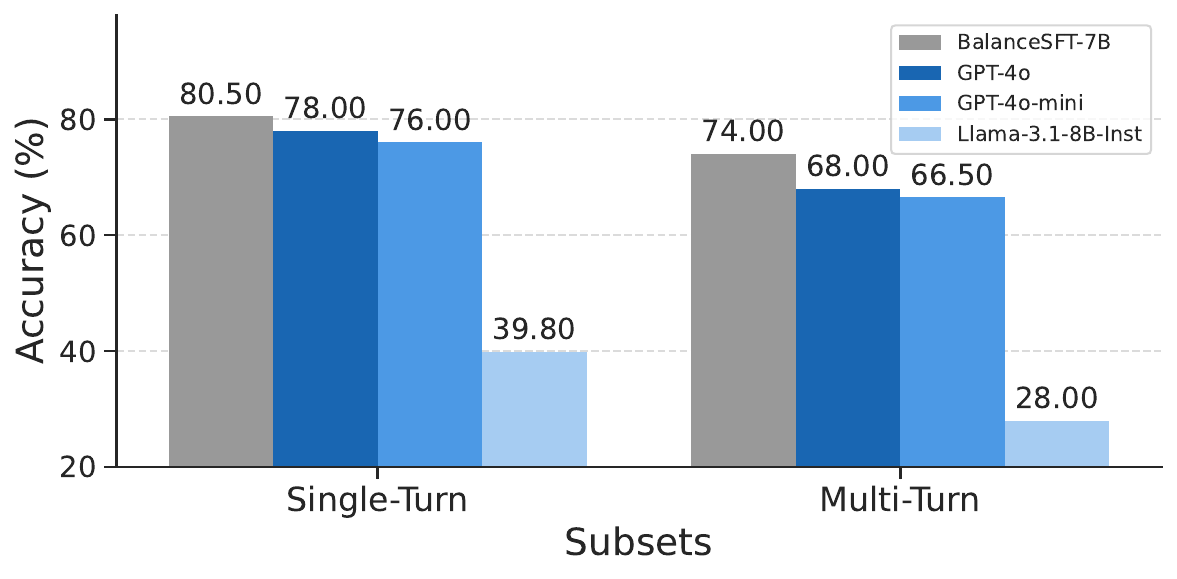}
        \caption{Performance on ACEBench.}
        \label{fig:left}
    \end{subfigure}
    \hfill
    \begin{subfigure}[b]{0.49\textwidth}
        \centering
        \includegraphics[width=\textwidth]{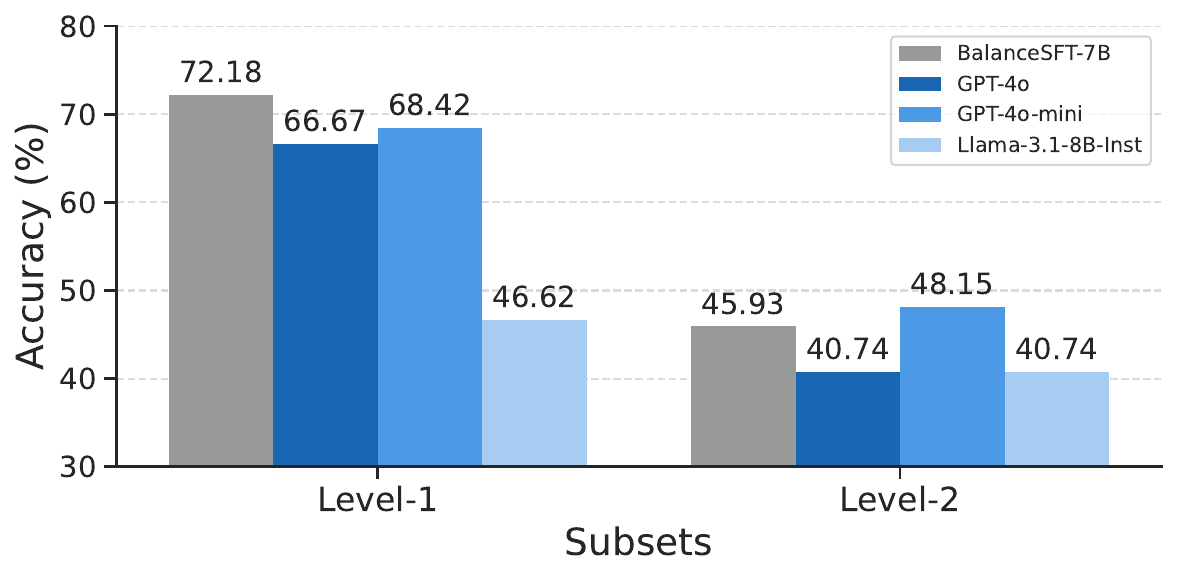}
        \caption{Performance on APIBank.}
        \label{fig:right}
    \end{subfigure}
    \caption{Performance comparison on ACEBench and APIBank benchmarks using official evaluation scripts, reported as accuracy (\%). }
    \label{tab:model_performance}
\end{figure*}

\noindent\textbf{Results on Other Benchmarks.}  Figure~\ref{tab:model_performance} presents the performance of BalanceSFT-7B on single-turn and multi-turn tasks from ACEBench~\cite{chen2025acebench}, the Level-1 (Call) and Level-2 (Retrieve+Call) subsets from APIBank~\cite{li2023api}.
On ACEBench, BalanceSFT-7B achieves the highest accuracy of 80.50 in the single-turn task, outperforming GPT-4o, which scores 78.00. In the multi-turn task, it also attains the best result with an accuracy of 74.00, which is 6 percentage points higher than GPT-4o (68.00). On APIBank, BalanceSFT-7B maintains strong generalization, ranking first and second on the Level-1 and Level-2 subsets. These results confirm that the model trained with BalanceSFT exhibits robust and consistent performance across diverse tool-calling benchmarks.
\subsection{Ablation Study}
In the ablation study, all experiments are conducted under consistent configurations to ensure fair and controlled comparisons. The basic SFT method serves as the baseline to assess the contribution of each proposed component. To further examine the generalization of our proposed method framework, we conduct experiments on two base models: Qwen2.5-Coder-Inst and Llama3.2-3B-Inst.


\noindent\textbf{Effectiveness of the Designed Components.}
Our ablation analysis, presented in Table \ref{tab:ablation}, systematically investigates the effect of the two core modules—Self-adjusted Signal Balancing (SSB) loss and Hard Data Re-sampling (HDR)—on both single-turn and multi-turn performance within the BFCL benchmark.


For Qwen2.5-Coder-Inst, the SFT trained model achieves 81.73 and 38.25 in Single-Turn and Multi-Turn, respectively. Individual activation of SSB yields improvements up to 82.20 (Single-Turn) and 41.62 (Multi-Turn). In contrast, applying SFT to HDR resampled data provides larger gains, boosting Single-Turn by 2.12 and dramatically increasing Multi-Turn by 4.87, suggesting that rare edge samples are critical. Crucially, the combination of both SSB and HDR resulted in synergistic gains, achieving the highest performance of 84.00 (Single-Turn) and 47.00 (Multi-Turn). These findings highlight the significant contribution of both modules to improving model robustness and accuracy. 

A similar pattern can be observed for Llama-3.2-3B-Inst, showing consistent trends, which further validates the generality of our approach. These results again validate that both SSB and HDR individually enhance performance, especially in the Multi-Turn scenario. Notably, their combination provides a complementary and superior method for optimizing tool-use capabilities across different model architectures, i.e., yielding more hard edge samples (through HDR) and enabling the model to learn more effectively (through SSB). 

\begin{table}[t]
\centering
\renewcommand{\arraystretch}{0.9}
\setlength{\tabcolsep}{4pt}
\small
\begin{tabular}{ccccc}
\toprule
\multirow{2}{*}{Models} & \multicolumn{2}{c}{Modules} & \multirow{2}{*}{\makecell{Single\\Turn}} & \multirow{2}{*}{\makecell{Multi\\Turn}} \\
\cmidrule(lr){2-3}
& SSB & HDR & &  \\
\midrule
Qwen2.5-Coder-Inst+SFT & \xmark & \xmark &81.73&38.25  \\
&  \cmark & \xmark & 82.20 & 41.62  \\
& \xmark& \cmark & 83.85 & 43.12  \\
&  \cmark & \cmark & 84.00 & 47.00  \\
\midrule
Llama-3.2-3B-Inst+SFT & \xmark & \xmark &73.67&33.25  \\
&  \cmark & \xmark & 76.82 & 34.62  \\
& \xmark& \cmark & 78.28 & 35.75  \\
&  \cmark & \cmark & 78.99 & 36.12  \\
\bottomrule
\end{tabular}
\caption{Performance comparison of Qwen2.5-Coder-Inst and Llama-3.2-3B-Inst, across different components of modules (SSB and HDR) in Single-Turn and Multi-Turn on BFCL.}
\vspace{-5mm} 
\label{tab:ablation}
\end{table}

\begin{figure*}[!t]
    \centering
    \begin{subfigure}[t]{0.325\textwidth}
        \centering
        \includegraphics[width=\textwidth]{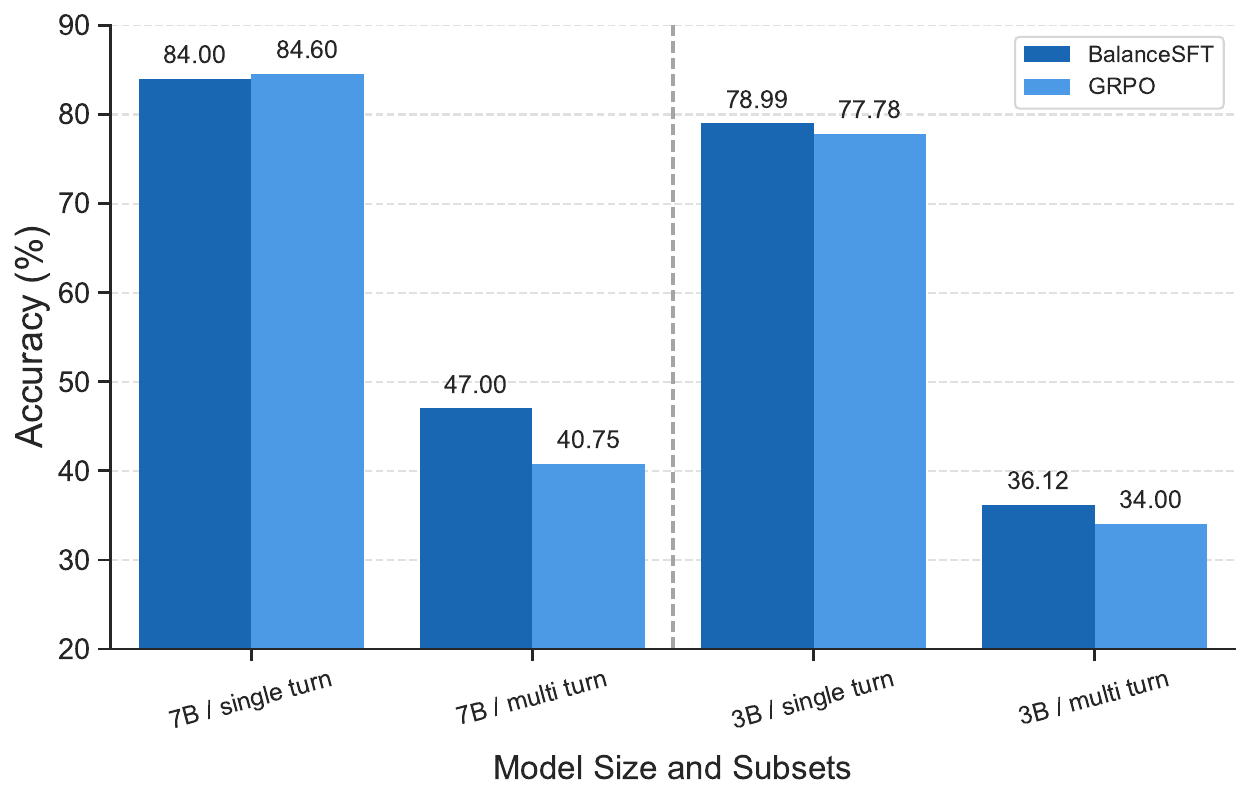}
        \caption{Compare with GRPO.}
        \label{fig:left}
    \end{subfigure}
    \hfill
    \begin{subfigure}[t]{0.325\textwidth}
        \centering
        \includegraphics[width=\textwidth]{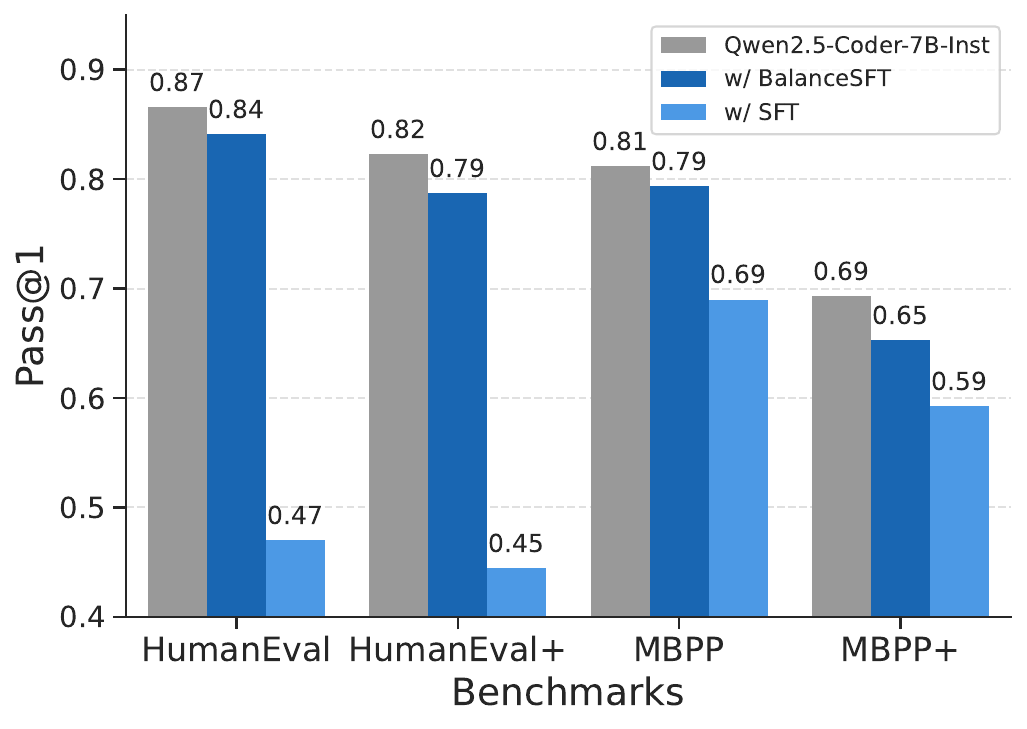}
        \caption{Performance On Code Evaluations.}
        \label{fig:mid}
    \end{subfigure}
    \hfill
    \begin{subfigure}[t]{0.325\textwidth}
        \centering
        \includegraphics[width=\textwidth]{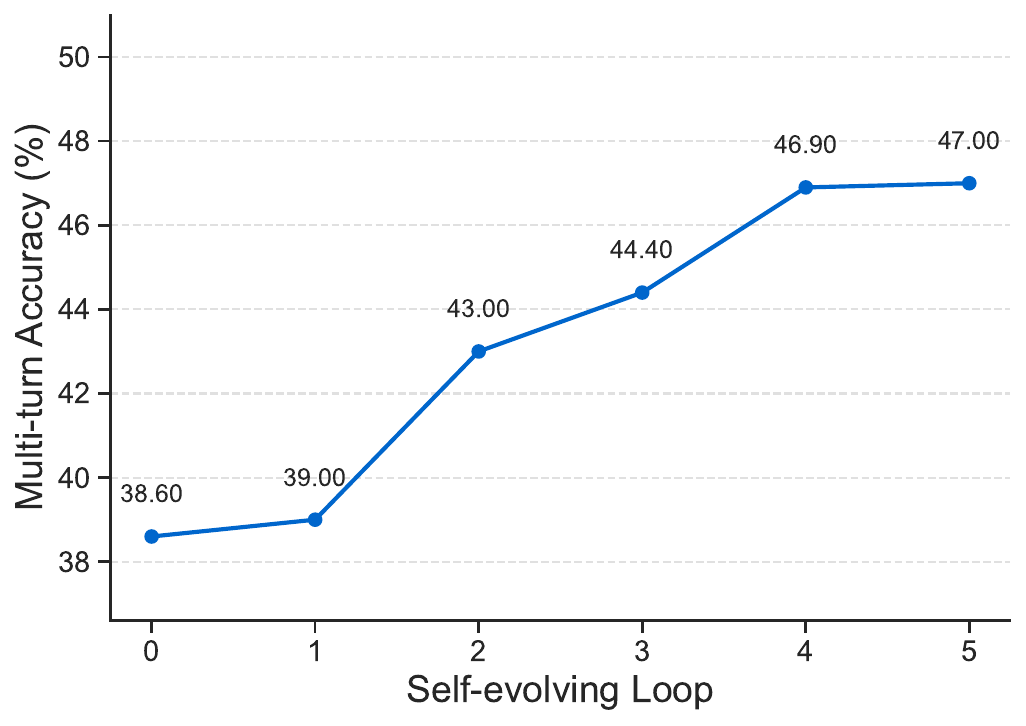}
        \caption{Performance at Different Loop stages.}
        \label{fig:right}
    \end{subfigure}
    \caption{\textbf{More experiments.} (a) Performance of two series models trained by GRPO and BalanceSFT on BFCL. (b) Performance of BalanceSFT and SFT models on HumanEval and MBPP (including HumanEval+ and MBPP+) compared with that of the code pre-trained model (Qwen2.5-Coder-7B-Inst). (c) Multi-Turn performance of BalanceSFT-7B at different stages in Self-evolving.}
    \label{code}
    \vspace{-5mm} 
\end{figure*}
\subsection{Further Analysis}

\noindent\textbf{Comparison with Reinforcement Learning.} To further validate the effectiveness of the proposed BalanceSFT method, we compare it with Group Relative Policy Optimization (GRPO)~\cite{guo2025deepseek} method on BFCL shown in Table~\ref{fig:left}, the 7B and 3B models are trained upon Qwen2.5-Coder-7B-Inst and Llama-3.2-3B-Inst~\cite{shao2024deepseekmath}. We use the hyperparameter settings and reward design from the official GRPO baseline described in Appendix~\ref{RLTrainingDetails}. For Qwen2.5-Coder-7B-Inst, the GRPO method achieves a higher score in the Single-Turn task (84.60 vs. 84.00), but the BalanceSFT approach outperforms it significantly in the Multi-Turn task (47.00 vs. 40.75). This suggests that while GRPO maintains an edge in simpler, Single-Turn interactions, the BalanceSFT method better enhances the model's capability to handle more complex, Multi-Turn dialogues. A similar trend is observed with Llama-3.2-3B-Inst, where BalanceSFT outperforms GRPO in both tasks, marginally in Single-Turn (78.99 vs. 77.78) and more notably in Multi-Turn (36.12 vs. 34.00). The results indicate that the proposed BalanceSFT approach provides a more balanced performance boost, particularly excelling in the more challenging Multi-Turn setting compared to the GRPO method, thereby validating its advantage in enhancing model robustness across different interaction complexities.

\noindent\textbf{Effectiveness of HDR Self-Evolving Loop}. As shown in Figure~\ref{fig:right}, the model's ability in Multi-Turn function call grows with the Self-evolving Loop until convergence, providing strong evidence that the proposed HDR can effectively and continuously sample complex edge samples, driving the model performance toward a high-accuracy.

This improvement arises from the HDR Self-Evolving Loop, which forms a self-reinforcing training cycle. HDR leverages model errors to create harder edge samples that target its weak points, forming a focused feedback loop for continual improvement. Through iterative refinement, the model progressively enhances its accuracy, robustness across complex, multi-step tasks.

\noindent\textbf{Effectiveness of Alleviating Catastrophic Forgetting.}
To assess how BalanceSFT impacts the coder model’s original coding capabilities, we evaluate our model on widely recognized code generation benchmarks in Figure~\ref{fig:mid}, HumanEval and MBPP (including their plus variants)~\cite{austin2021program,chen2021evaluating,yu2024humaneval}. The results, presented in Table~\ref{fig:right}, demonstrate a crucial advantage of our method. Specifically, BalanceSFT-7B maintains strong performance across all benchmarks, with pass@1 scores only marginally lower (within 4\%) compared to the base Qwen2.5-Coder-Inst model. For instance, on HumanEval, BalanceSFT achieves 0.841 compared to the base model's 0.866, and on MBPP, it reaches 0.794 versus 0.812. This minimal performance degradation stands in stark contrast to models fine-tuned using standard SFT. The SFT method shows dramatic drops in pass@1 scores on all the benchmarks, notably falling from 0.866 to 0.470 on HumanEval and from 0.823 to 0.445 on HumanEval+. 

These results indicate that BalanceSFT effectively mitigates catastrophic forgetting, preserving the model’s original code capabilities across benchmarks. This is because the SSB loss dynamically rebalances the training signal, focusing updates on the concise, yet crucial function call tokens. By concentrating learning on these key tokens, the model avoids overfitting to lengthy reasoning and maintains its original coding capabilities.

\section{Conclusion}
In this work, we identify two critical yet overlooked challenges in standard SFT for function calling: the imbalance in training signals between lengthy reasoning and concise function calls, and the scarcity of hard training data. To address these issues, we propose BalanceSFT, a novel framework that integrates a Self-adjusted Signal Balancing (SSB) loss and a Hard Data Re-sampling (HDR) strategy. The SSB loss dynamically recalibrates the learning focus between reasoning and execution, while HDR iteratively enriches the training set with high-quality complex examples. Extensive experiments demonstrate that BalanceSFT enables a 7B model to achieve function calling performance competitive with state-of-the-art giants like GPT-4o on BFCL and other benchmarks. Our work provides a simple yet effective pathway toward more robust and generalizable LLM-based tool use.


\clearpage

\section*{Limitations}
Despite its strong performance, BalanceSFT has two major limitations. First, the effectiveness of the SSB loss depends on the initial quality and structure of the CoT data, which may limit its applicability in domains where high-quality reasoning annotations are scarce. Second, while HDR improves data hardness, it relies on an ensemble of LLM judges for quality control, which introduces computational cost and potential bias from the judge models.

\bibliography{main}

@article{sheng2024hybridflow,
  title   = {HybridFlow: A Flexible and Efficient RLHF Framework},
  author  = {Guangming Sheng and Chi Zhang and Zilingfeng Ye and Xibin Wu and Wang Zhang and Ru Zhang and Yanghua Peng and Haibin Lin and Chuan Wu},
  year    = {2024},
  journal = {arXiv preprint arXiv: 2409.19256}
}

@inproceedings{kwon2023efficient,
  title={Efficient Memory Management for Large Language Model Serving with PagedAttention},
  author={Woosuk Kwon and Zhuohan Li and Siyuan Zhuang and Ying Sheng and Lianmin Zheng and Cody Hao Yu and Joseph E. Gonzalez and Hao Zhang and Ion Stoica},
  booktitle={Proceedings of the ACM SIGOPS 29th Symposium on Operating Systems Principles},
  year={2023}
}

@techreport{openai2025gpt,
  title={GPT-5 System Card},
  author={OpenAI},
  year={2025},
  month={August},
  institution={OpenAI},
  type={Technical report}
}

@inproceedings{yang2025alleviating,
  title={Alleviating the Fear of Losing Alignment in LLM Fine-tuning},
  author={Yang, Kang and Tao, Guanhong and Chen, Xun and Xu, Jun},
  booktitle={2025 IEEE Symposium on Security and Privacy (SP)},
  pages={2152--2170},
  year={2025},
  organization={IEEE}
}

@inproceedings{lobo2025impact,
  title={On the Impact of Fine-Tuning on Chain-of-Thought Reasoning},
  author={Lobo, Elita and Agarwal, Chirag and Lakkaraju, Himabindu},
  booktitle={Proceedings of the 2025 Conference of the Nations of the Americas Chapter of the Association for Computational Linguistics: Human Language Technologies (Volume 1: Long Papers)},
  pages={11679--11698},
  year={2025}
}

@article{chen2025reasoning,
  title={Reasoning Models Don't Always Say What They Think},
  author={Chen, Yanda and Benton, Joe and Radhakrishnan, Ansh and Uesato, Jonathan and Denison, Carson and Schulman, John and Somani, Arushi and Hase, Peter and Wagner, Misha and Roger, Fabien and others},
  journal={arXiv preprint arXiv:2505.05410},
  year={2025}
}

@article{agarwal2024faithfulness,
  title={Faithfulness vs. plausibility: On the (un) reliability of explanations from large language models},
  author={Agarwal, Chirag and Tanneru, Sree Harsha and Lakkaraju, Himabindu},
  journal={arXiv preprint arXiv:2402.04614},
  year={2024}
}

@article{dong2025tool,
  title={Tool-Star: Empowering LLM-Brained Multi-Tool Reasoner via Reinforcement Learning},
  author={Dong, Guanting and Chen, Yifei and Li, Xiaoxi and Jin, Jiajie and Qian, Hongjin and Zhu, Yutao and Mao, Hangyu and Zhou, Guorui and Dou, Zhicheng and Wen, Ji-Rong},
  journal={arXiv preprint arXiv:2505.16410},
  year={2025}
}

@inproceedings{wu2025joint,
  title={A joint optimization framework for enhancing efficiency of tool utilization in LLM agents},
  author={Wu, Bin and Meij, Edgar and Yilmaz, Emine},
  booktitle={Findings of the Association for Computational Linguistics: ACL 2025},
  pages={22361--22373},
  year={2025}
}

@article{shen2024llm,
  title={Llm with tools: A survey},
  author={Shen, Zhuocheng},
  journal={arXiv preprint arXiv:2409.18807},
  year={2024}
}

@inproceedings{li2025review,
  title={A review of prominent paradigms for llm-based agents: Tool use, planning (including rag), and feedback learning},
  author={Li, Xinzhe},
  booktitle={Proceedings of the 31st International Conference on Computational Linguistics},
  pages={9760--9779},
  year={2025}
}

@inproceedings{chen2024towards,
  title={Towards tool use alignment of large language models},
  author={Chen, Zhi-Yuan and Shen, Shiqi and Shen, Guangyao and Zhi, Gong and Chen, Xu and Lin, Yankai},
  booktitle={Proceedings of the 2024 Conference on Empirical Methods in Natural Language Processing},
  pages={1382--1400},
  year={2024}
}

@inproceedings{kokane2025toolscan,
  title={ToolScan: A Benchmark For Characterizing Errors In Tool-Use LLMs},
  author={Kokane, Shirley and Zhu, Ming and Awalgaonkar, Tulika Manoj and Zhang, Jianguo and Prabhakar, Akshara and Hoang, Thai Quoc and Liu, Zuxin and RN, Rithesh and Yang, Liangwei and Yao, Weiran and others},
  booktitle={ICLR 2025 Workshop on Building Trust in Language Models and Applications}
}

@article{lu2024toolsandbox,
  title={Toolsandbox: A stateful, conversational, interactive evaluation benchmark for llm tool use capabilities},
  author={Lu, Jiarui and Holleis, Thomas and Zhang, Yizhe and Aumayer, Bernhard and Nan, Feng and Bai, Felix and Ma, Shuang and Ma, Shen and Li, Mengyu and Yin, Guoli and others},
  journal={arXiv preprint arXiv:2408.04682},
  year={2024}
}

@article{yu2024humaneval,
  title={Humaneval pro and mbpp pro: Evaluating large language models on self-invoking code generation},
  author={Yu, Zhaojian and Zhao, Yilun and Cohan, Arman and Zhang, Xiao-Ping},
  journal={arXiv preprint arXiv:2412.21199},
  year={2024}
}

@inproceedings{qu2025latent,
  title={Latent reward: Llm-empowered credit assignment in episodic reinforcement learning},
  author={Qu, Yun and Jiang, Yuhang and Wang, Boyuan and Mao, Yixiu and Wang, Cheems and Liu, Chang and Ji, Xiangyang},
  booktitle={Proceedings of the AAAI Conference on Artificial Intelligence},
  volume={39},
  number={19},
  pages={20095--20103},
  year={2025}
}

@article{yang2025qwen3,
  title={Qwen3 technical report},
  author={Yang, An and Li, Anfeng and Yang, Baosong and Zhang, Beichen and Hui, Binyuan and Zheng, Bo and Yu, Bowen and Gao, Chang and Huang, Chengen and Lv, Chenxu and others},
  journal={arXiv preprint arXiv:2505.09388},
  year={2025}
}

@article{zeng2025toolaceacer,
  title={ToolACE-R: Tool Learning with Adaptive Self-Refinement},
  author={Zeng, Xingshan and Liu, Weiwen and Huang, Xu and Wang, Zezhong and Wang, Lingzhi and Li, Liangyou and Wang, Yasheng and Shang, Lifeng and Jiang, Xin and Tang, Ruiming and others},
  journal={arXiv preprint arXiv:2504.01400},
  year={2025}
}

@article{zheng2023llm,
  title={Llm-funcmapper: Function identification for interpreting complex clauses in building codes via llm},
  author={Zheng, Zhe and Chen, Ke-Yin and Cao, Xin-Yu and Lu, Xin-Zheng and Lin, Jia-Rui},
  journal={arXiv preprint arXiv:2308.08728},
  year={2023}
}

@article{wang2023describe,
  title={Describe, explain, plan and select: interactive planning with llms enables open-world multi-task agents},
  author={Wang, Zihao and Cai, Shaofei and Chen, Guanzhou and Liu, Anji and Ma, Xiaojian Shawn and Liang, Yitao},
  journal={Advances in Neural Information Processing Systems},
  volume={36},
  pages={34153--34189},
  year={2023}
}

@article{ju2024mitigating,
  title={Mitigating training imbalance in llm fine-tuning via selective parameter merging},
  author={Ju, Yiming and Ni, Ziyi and Xing, Xingrun and Zeng, Zhixiong and Fan, Siqi and Zhang, Zheng and others},
  journal={arXiv preprint arXiv:2410.03743},
  year={2024}
}

@misc{Open-Agentic-tool-use,
    title = {Open-Agentic-tool-use},
    url = {https://github.com/BingguangHao/Open-Agentic-tool-use},
    author = {{Aworld team, inclusionAI}},
    year = {2025}
}

@article{li2023api,
  title={Api-bank: A comprehensive benchmark for tool-augmented llms},
  author={Li, Minghao and Zhao, Yingxiu and Yu, Bowen and Song, Feifan and Li, Hangyu and Yu, Haiyang and Li, Zhoujun and Huang, Fei and Li, Yongbin},
  journal={arXiv preprint arXiv:2304.08244},
  year={2023}
}

@article{chen2025acebench,
  title={ACEBench: Who Wins the Match Point in Tool Learning?},
  author={Chen, Chen and Hao, Xinlong and Liu, Weiwen and Huang, Xu and Zeng, Xingshan and Yu, Shuai and Li, Dexun and Wang, Shuai and Gan, Weinan and Huang, Yuefeng and others},
  journal={arXiv e-prints},
  pages={arXiv--2501},
  year={2025}
}

@inproceedings{Bitagent,
  title={Bitagent},
year={2024}
}

@inproceedings{grok4,
  title={Grok 4},
  author={xAI},
year={2025}
}

@inproceedings{TheC3,
  title={The Claude 3 Model Family: Opus, Sonnet, Haiku},
  author={Claude},
}

@article{hurst2024gpt,
  title={Gpt-4o system card},
  author={Hurst, Aaron and Lerer, Adam and Goucher, Adam P and Perelman, Adam and Ramesh, Aditya and Clark, Aidan and Ostrow, AJ and Welihinda, Akila and Hayes, Alan and Radford, Alec and others},
  journal={arXiv preprint arXiv:2410.21276},
  year={2024}
}

@article{comanici2025gemini,
  title={Gemini 2.5: Pushing the frontier with advanced reasoning, multimodality, long context, and next generation agentic capabilities},
  author={Comanici, Gheorghe and Bieber, Eric and Schaekermann, Mike and Pasupat, Ice and Sachdeva, Noveen and Dhillon, Inderjit and Blistein, Marcel and Ram, Ori and Zhang, Dan and Rosen, Evan and others},
  journal={arXiv preprint arXiv:2507.06261},
  year={2025}
}

@article{zeng2025toolace,
  title={ToolACE-MT: Non-Autoregressive Generation for Agentic Multi-Turn Interaction},
  author={Zeng, Xingshan and Liu, Weiwen and Wang, Lingzhi and Li, Liangyou and Mi, Fei and Wang, Yasheng and Shang, Lifeng and Jiang, Xin and Liu, Qun},
  journal={arXiv preprint arXiv:2508.12685},
  year={2025}
}

@article{shao2024deepseekmath,
  title={Deepseekmath: Pushing the limits of mathematical reasoning in open language models},
  author={Shao, Zhihong and Wang, Peiyi and Zhu, Qihao and Xu, Runxin and Song, Junxiao and Bi, Xiao and Zhang, Haowei and Zhang, Mingchuan and Li, YK and Wu, Y and others},
  journal={arXiv preprint arXiv:2402.03300},
  year={2024}
}

@inproceedings{wang2024gta,
  title={GTA: a benchmark for general tool agents},
  author={Wang, Jize and Zerun, Ma and Li, Yining and Zhang, Songyang and Chen, Cailian and Chen, Kai and Le, Xinyi},
  booktitle={The Thirty-eight Conference on Neural Information Processing Systems Datasets and Benchmarks Track},
  year={2024}
}

@article{kavathekar2025small,
  title={Small Models, Big Tasks: An Exploratory Empirical Study on Small Language Models for Function Calling},
  author={Kavathekar, Ishan and Donakanti, Raghav and Kumaraguru, Ponnurangam and Vaidhyanathan, Karthik},
  journal={arXiv preprint arXiv:2504.19277},
  year={2025}
}

@inproceedings{lightman2023let,
  title={Let's verify step by step},
  author={Lightman, Hunter and Kosaraju, Vineet and Burda, Yuri and Edwards, Harrison and Baker, Bowen and Lee, Teddy and Leike, Jan and Schulman, John and Sutskever, Ilya and Cobbe, Karl},
  booktitle={The Twelfth International Conference on Learning Representations},
  year={2023}
}

@article{ouyang2022training,
  title={Training language models to follow instructions with human feedback},
  author={Ouyang, Long and Wu, Jeffrey and Jiang, Xu and Almeida, Diogo and Wainwright, Carroll and Mishkin, Pamela and Zhang, Chong and Agarwal, Sandhini and Slama, Katarina and Ray, Alex and others},
  journal={Advances in neural information processing systems},
  volume={35},
  pages={27730--27744},
  year={2022}
}

@article{bai2023qwen,
  title={Qwen technical report},
  author={Bai, Jinze and Bai, Shuai and Chu, Yunfei and Cui, Zeyu and Dang, Kai and Deng, Xiaodong and Fan, Yang and Ge, Wenbin and Han, Yu and Huang, Fei and others},
  journal={arXiv preprint arXiv:2309.16609},
  year={2023}
}

@article{qwq,
  title={QwQ: Reflect deeply on the boundaries of the unknown},
  author={Qwen Team},
  year={2024}
}

@article{zhong2024evaluation,
  title={Evaluation of openai o1: Opportunities and challenges of agi},
  author={Zhong, Tianyang and Liu, Zhengliang and Pan, Yi and Zhang, Yutong and Zhou, Yifan and Liang, Shizhe and Wu, Zihao and Lyu, Yanjun and Shu, Peng and Yu, Xiaowei and others},
  journal={arXiv preprint arXiv:2409.18486},
  year={2024}
}

@article{team2025kimi,
  title={Kimi k2: Open agentic intelligence},
  author={Team, Kimi and Bai, Yifan and Bao, Yiping and Chen, Guanduo and Chen, Jiahao and Chen, Ningxin and Chen, Ruijue and Chen, Yanru and Chen, Yuankun and Chen, Yutian and others},
  journal={arXiv preprint arXiv:2507.20534},
  year={2025}
}

@inproceedings{wang2024executable,
  title={Executable code actions elicit better llm agents},
  author={Wang, Xingyao and Chen, Yangyi and Yuan, Lifan and Zhang, Yizhe and Li, Yunzhu and Peng, Hao and Ji, Heng},
  booktitle={Forty-first International Conference on Machine Learning},
  year={2024}
}

@article{singh2024llm,
  title={An llm-tool compiler for fused parallel function calling},
  author={Singh, Simranjit and Karatzas, Andreas and Fore, Michael and Anagnostopoulos, Iraklis and Stamoulis, Dimitrios},
  journal={arXiv preprint arXiv:2405.17438},
  year={2024}
}

@article{wang2025function,
  title={Function Calling in Large Language Models: Industrial Practices, Challenges, and Future Directions},
  author={WANG, MAOLIN and ZHANG, YINGYI and PENG, CUNYIN and CHEN, YICHENG and ZHOU, WEI and GU, JINJIE and ZHUANG, CHENYI and GUO, RUOCHENG and YU, BOWEN and WANG, WANYU and others},
  year={2025}
}

@article{prabhakar2025apigen,
  title={Apigen-mt: Agentic pipeline for multi-turn data generation via simulated agent-human interplay},
  author={Prabhakar, Akshara and Liu, Zuxin and Yao, Weiran and Zhang, Jianguo and Zhu, Ming and Wang, Shiyu and Liu, Zhiwei and Awalgaonkar, Tulika and Chen, Haolin and Hoang, Thai and others},
  journal={arXiv preprint arXiv:2504.03601},
  year={2025}
}

@article{zheng2024llamafactory,
  title={Llamafactory: Unified efficient fine-tuning of 100+ language models},
  author={Zheng, Yaowei and Zhang, Richong and Zhang, Junhao and Ye, Yanhan and Luo, Zheyan and Feng, Zhangchi and Ma, Yongqiang},
  journal={arXiv preprint arXiv:2403.13372},
  year={2024}
}

@article{hui2024qwen2,
  title={Qwen2. 5-coder technical report},
  author={Hui, Binyuan and Yang, Jian and Cui, Zeyu and Yang, Jiaxi and Liu, Dayiheng and Zhang, Lei and Liu, Tianyu and Zhang, Jiajun and Yu, Bowen and Lu, Keming and others},
  journal={arXiv preprint arXiv:2409.12186},
  year={2024}
}

@article{austin2021program,
  title={Program synthesis with large language models},
  author={Austin, Jacob and Odena, Augustus and Nye, Maxwell and Bosma, Maarten and Michalewski, Henryk and Dohan, David and Jiang, Ellen and Cai, Carrie and Terry, Michael and Le, Quoc and others},
  journal={arXiv preprint arXiv:2108.07732},
  year={2021}
}

@article{chen2021evaluating,
  title={Evaluating large language models trained on code},
  author={Chen, Mark and Tworek, Jerry and Jun, Heewoo and Yuan, Qiming and Pinto, Henrique Ponde De Oliveira and Kaplan, Jared and Edwards, Harri and Burda, Yuri and Joseph, Nicholas and Brockman, Greg and others},
  journal={arXiv preprint arXiv:2107.03374},
  year={2021}
}

@article{patil2024gorilla,
  title={Gorilla: Large language model connected with massive apis},
  author={Patil, Shishir G and Zhang, Tianjun and Wang, Xin and Gonzalez, Joseph E},
  journal={Advances in Neural Information Processing Systems},
  volume={37},
  pages={126544--126565},
  year={2024}
}

@article{zhang2025nemotron,
    year = {2025},
    title = {Nemotron-Research-Tool-N1: Tool-Using Language Models with Reinforced Reasoning},
    author = {Zhang, Shaokun and Dong, Yi and Zhang, Jieyu and Kautz, Jan and Catanzaro, Bryan and Tao, Andrew and Wu, Qingyun and Yu, Zhiding and Liu, Guilin},
    journal = {arXiv preprint arXiv:2505.00024}}

@article{qian2025toolrl,
    year = {2025},
    title = {Toolrl: Reward is all tool learning needs},
    author = {Qian, Cheng and Acikgoz, Emre Can and He, Qi and Wang, Hongru and Chen, Xiusi and Hakkani-T{\"u}r, Dilek and Tur, Gokhan and Ji, Heng},
    journal = {arXiv preprint arXiv:2504.13958}}

@article{guo2025deepseek,
    year = {2025},
    title = {Deepseek-r1: Incentivizing reasoning capability in llms via reinforcement learning},
    author = {Guo, Daya and Yang, Dejian and Zhang, Haowei and Song, Junxiao and Zhang, Ruoyu and Xu, Runxin and Zhu, Qihao and Ma, Shirong and Wang, Peiyi and Bi, Xiao and others},
    journal = {arXiv preprint arXiv:2501.12948}}

@article{achiam2023gpt,
  title={Gpt-4 technical report},
  author={Achiam, Josh and Adler, Steven and Agarwal, Sandhini and Ahmad, Lama and Akkaya, Ilge and Aleman, Florencia Leoni and Almeida, Diogo and Altenschmidt, Janko and Altman, Sam and Anadkat, Shyamal and others},
  journal={arXiv preprint arXiv:2303.08774},
  year={2023}}

@article{touvron2023llama,
  title={Llama: Open and efficient foundation language models},
  author={Touvron, Hugo and Lavril, Thibaut and Izacard, Gautier and Martinet, Xavier and Lachaux, Marie-Anne and Lacroix, Timoth{\'e}e and Rozi{\`e}re, Baptiste and Goyal, Naman and Hambro, Eric and Azhar, Faisal and others},
  journal={arXiv preprint arXiv:2302.13971},
  year={2023}}

@article{zhang2024xlam,
  title={xlam: A family of large action models to empower ai agent systems},
  author={Zhang, Jianguo and Lan, Tian and Zhu, Ming and Liu, Zuxin and Hoang, Thai and Kokane, Shirley and Yao, Weiran and Tan, Juntao and Prabhakar, Akshara and Chen, Haolin and others},
  journal={arXiv preprint arXiv:2409.03215},
  year={2024}
}

@article{2025InSpire,
  title={InSpire: Vision-Language-Action Models with Intrinsic Spatial Reasoning},
  author={ Zhang, Ji  and  Wu, Shihan  and  Luo, Xu  and  Wu, Hao  and  Gao, Lianli  and  Shen, Heng Tao  and  Song, Jingkuan },
  journal={arXiv preprint arXiv:2505.13888},
  year={2025},
}

\appendix
\section{Dataset}

\begin{figure*}[h]
    \centering
    \includegraphics[width=\textwidth]{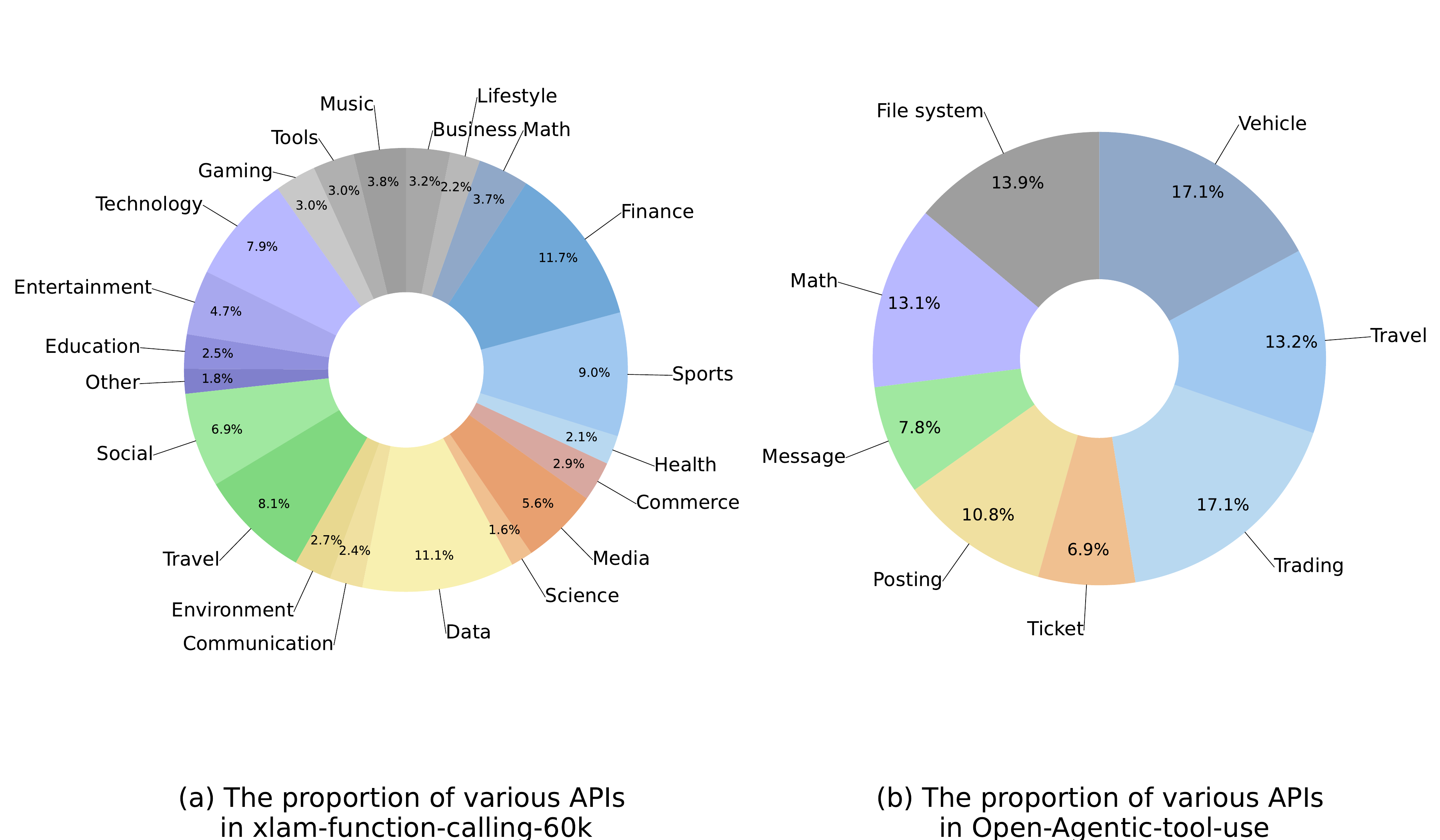}
    \caption{API categories distribution of xlam-function-calling-60k and Open-Agentic-tool-use. Different colors represent the distribution of different API categories.}
    \label{fig:APIategories}
\end{figure*}

\begin{figure*}[h]
\centering
\begin{tcolorbox}[
    colback=gray!5!white,
    title={Error of xlam Dataset},
    boxrule=0.3mm,
    width=0.95\textwidth,
    arc=3mm,
    auto outer arc=true,
]
\textbf{Query:} 
What are the daily weather forecasts for Chicago and Toronto for the next 7 days?\vspace{0.8em}
\textbf{Tools:}
\begin{small}
\begin{lstlisting}[basicstyle=\ttfamily\footnotesize,backgroundcolor=\color{gray!5!white},frame=none,breaklines=true,showstringspaces=false]
[
  {
    "name": "get_forecastdata_by_lat_lon",
    "description": "Fetches a 14-day weather forecast for a given location specified by latitude and longitude.",
    "parameters": {
      "lat": {"type": "int", "default": "51.5"},
      "lon": {"type": "int", "default": "-0.6"},
      "lang": {"type": "str", "default": "en"}
    }
  },
  {
    "name": "location_info",
    "description": "Retrieve metadata for a specified location.",
    "parameters": {
      "location": {"type": "str", "default": "101275339"}
    }
  },
  {
    "name": "forecast_weather_api",
    "description": "Fetches weather forecast and alerts from API.",
    "parameters": {
      "q": {"type": "str", "default": "London"},
      "days": {"type": "int", "default": "3"},
      "dt": {"type": "str", "default": ""},
      "lang": {"type": "str", "default": ""}
    }
  },
  {
    "name": "daily",
    "description": "Fetches daily weather forecast for a given location.",
    "parameters": {
      "location": {"type": "str", "default": "102643743"},
      "alt": {"type": "int", "default": "0"},
      "dataset": {"type": "str", "default": "full"},
      "tempunit": {"type": "str", "default": "C"},
      "windunit": {"type": "str", "default": "MS"},
      "periods": {"type": "int", "default": "8"}
    }
  }
]
\end{lstlisting}
\end{small}
\textbf{Answer:}
\begin{small}
\ttfamily
\newline[\newline
\hspace*{1em}\{\newline
\hspace*{2em}"name": "{\colorbox{red!30}{\hspace{0.6em}}}forecast\_weather\_api",\newline
\hspace*{2em}"arguments": \{\newline
\hspace*{3em}"q": "Chicago",\newline
\hspace*{3em}"days": 7 \}
\newline \hspace*{1em}\}
\newline \hspace*{1em}\{\newline
\hspace*{2em}"name": "forecast\_weather\_api",\newline
\hspace*{2em}"arguments": \{\newline
\hspace*{3em}"q": "Toronto",\newline
\hspace*{3em}"days": 7 \}
\newline\hspace*{1em}\}\newline
]
\end{small}
\end{tcolorbox}
\caption{A Example for error in xlam dataset}
\label{fig:badcase_weather}
\end{figure*}

\label{Dataset}


\subsection{Data Analysis}

Here, we split each turn of multi-turn conversations and count them together with Single-Turn. By using Qwen2.5-Coder-7B-Instruct's tokenizer for token calculation, the average token length of function call of the entire dataset (including rejections) is 31.07, the average cot token length is 350.74. In the first data filtering, we obtain 85014 qualified simple data and 3178 difficult data for self-evolving, dropping 1808 samples.


\subsection{Error in Dataset}

As illustrated in Figure~\ref{fig:badcase_weather}, the xlam dataset contains an error in the formatting of function calls. Specifically, an extra space appears between the opening bracket and the function name ("[ forecast\_weather\_api(...)"), which causes the Abstract Syntax Tree (AST) parser to fail. This type of issue highlights the importance of performing a base quality check on the dataset to ensure correct syntax and reliable downstream processing. Ensuring proper formatting is essential for both automated function call execution and model evaluation.

\section{RL Training Details}
\label{RLTrainingDetails}

\begin{table*}[h]
\centering
\begin{tabular}{llllll}
\toprule
\textbf{Hyperparameter} & \textbf{Value} & \textbf{Hyperparameter} & \textbf{Value} & \textbf{Hyperparameter} & \textbf{Value} \\
\midrule
Batch Size & 1024 & Max Prompt Length & 8192 & Max Response Length & 20480 \\
Learning Rate & 1e-6 & Temperature & 0.7 & Epoch Number & 5 \\
KL Coefficient & 1e-3 & Entropy Coefficient & 1e-3 & Rollout Number & 8 \\
\bottomrule
\end{tabular}
\caption{The detailed hyperparameters used for RL training.}
\label{table:rl_hyperparameters}
\end{table*}

To further compare the methods, we select the reinforcement learning baseline method GRPO and conduct experiments by using both Qwen2.5-Coder-7B-Instruct and Llama-3.2-3B-Instruct. We use the open-source reinforcement learning library Verl~\cite{sheng2024hybridflow} for training; the key hyperparameters for training are shown in Table~\ref{table:rl_hyperparameters}. 

\section{More Results}

\subsection{Full BFCL Single-Turn Results}

We show the full single-turn results on the BFCL benchmark, where BalanceSFT attains superior performance with scores of 88.29 and 79.70 on non-live and live sections. The results surpass most similar scale models and achieve great enhancement on all the categories of live section, which indicates that our model has achieved excellent tool-calling capabilities under complex queries; the evaluation details are shown in Table~\ref{tab:single}.

\begin{table*}[thb]
\scriptsize
\centering
\begin{tabular}{@{}lcccccccccc@{}}
\toprule

& \multicolumn{5}{c}{\textbf{Non-Live}} & \multicolumn{5}{c}{\textbf{Live}}  \\
\cmidrule(lr){2-6} \cmidrule(lr){7-11}
\textbf{Models} &Overall &Simple & Multiple & Parallel & \begin{tabular}[c]{@{}c@{}}Parallel\\ Multiple\end{tabular} &Overall& Simple & Multiple & Parallel & \begin{tabular}[c]{@{}c@{}}Parallel\\ Multiple\end{tabular} \\
\midrule
\multicolumn{11}{c}{Close-Sourced Models} \\
\midrule
\textbf{GPT-4o-2024-11-20}&83.88	&76.50	&91.00	&90.00	&78.00	&70.54	&70.54	&70.75	&62.50	&66.67\\
\textbf{Gemini-2.5-Pro} &85.04	&68.67	&91.00	&91.50	&89.00	&63.95	&75.58	&60.40	&81.25	&83.33\\
\textbf{o3-2025-04-16}&39.98	&71.42	&88.50	&0.00	&0.00	&66.03	&69.77	&67.62	&0.00	&0.00	\\
\textbf{Amazon-Nova-Pro-v1:0}&85.25	&71.50	&93.50	&92.00	&84.00	&78.31	&80.62	&77.87	&87.50	&66.67\\
\textbf{Grok-4-0709}&85.21	&73.33	&92.50	&89.50	&85.50	&74.39	&77.91	&73.69	&68.75	&70.83\\
  \midrule
\multicolumn{11}{c}{Open-Sourced Models} \\
\midrule
\textbf{Moonshotai-Kimi-K2-Inst}&84.02	&79.08	&93.00	&76.00	&88.00	&77.57	&86.05	&75.78	&81.25	&62.50\\
\textbf{DeepSeek-R1-0528}&86.52	&71.58	&96.00	&91.00	&87.50	&77.65	&83.72	&76.16	&87.50	&70.83\\	
\textbf{Qwen3-235B-A22B-Inst-2507}&90.12	&80.50	&95.00	&95.50	&89.50	&76.61	&83.72	&75.02	&75.00	&70.83	\\
\textbf{Llama-4-Maverick-17B-128E-Inst-FP8}&88.15	&76.08	&95.00	&92.50	&89.00	&73.65	&84.50	&71.04	&75.00	&70.83\\
\textbf{Qwen3-8B}&88.60	&78.92	&95.00	&91.50	&89.00	&77.79	&80.23	&77.21	&81.25	&75.00\\	
\textbf{BitAgent-8B}&87.33	&76.33	&95.00	&93.00	&85.00	&76.09	&78.29	&75.59	&87.50	&66.67\\
\textbf{watt-tool-8B}&87.54	&76.17	&94.50	&94.00	&85.50	&75.87	&77.52	&75.59	&87.50	&62.50\\
\midrule
\multicolumn{11}{c}{Base Models} \\
\midrule
\textbf{Qwen2.5-Coder-7B-Instruct}&83.88&76.00&95.00&81.00&83.50&69.75&74.81&75.31&68.75&62.50\\
\textbf{Llama-3.2-3B-Instruct}&81.94	&71.25	&92.00	&87.00	&77.50	&57.74	&63.57	&57.26	&25.00	&37.50\\
\midrule
\multicolumn{11}{c}{Ours} \\
\midrule
\textbf{BalanceSFT-7B} & 88.29 &78.17&94.00&91.50&89.50&79.70&82.56&77.87&87.50&83.33 \\
\textbf{BalanceSFT-3B} & 83.83&73.33&90.50&88.00&83.50&74.14&74.42&71.70&75.00&79.17 \\
\bottomrule
\end{tabular}
\caption{The evaluation details on BFCL benchmark (last updated August 26, 2025), with all metrics calculated using the official script.}
\label{tab:single}
\end{table*}
\subsection{Results of Agentic Evaluation}
In addition to single-turn and multi-turn function call capabilities, BFCLv4 further evaluates the model's agentic capability, specifically, Web Search and Memory. We test our model on the corresponding test sets in Table~\ref{tab:bfclv4}, even though our base model has almost no agentic capability and we do not include any Web Search or Memory related data in the BalanceSFT training, our model still shows improvement in these two evaluation subsets, which indicates that our training and data selection are beneficial to the model's agentic ability.

\begin{table*}[h]
\centering
\begin{tabular}{@{}lccccccc@{}}
\toprule
&\multicolumn{3}{c}{\textbf{Web Search}}& \multicolumn{4}{c}{\textbf{Memory}}\\
\cmidrule(lr){2-4}\cmidrule(lr){5-8}
\textbf{Models}
&\multicolumn{1}{c}{\textit{Overall}} &
  \multicolumn{1}{c}{\textit{Base}} &
  \multicolumn{1}{c}{ \textit{No Snippet}} &
  \multicolumn{1}{c}{\textit{Overall}} &
  \multicolumn{1}{c}{\textit{KV}} &
  \multicolumn{1}{c}{\textit{Vector}}&
  \multicolumn{1}{c}{\textit{Recursive Sum}} \\ 
  \midrule
\textbf{Qwen2.5-Coder-7B-Instruct} &4.50	&4.00	&5.00	&2.37	&0.00&0.00	&7.10\\
\textbf{BalanceSFT-7B} &7.00	&9.00	&5.00	&8.17	&0.00&0.00	&24.52\\
\bottomrule
\end{tabular}
\caption{\label{tab:bfclv4} Performance of model's agentic ability.}
\end{table*}


\subsection{Results on ACEBench}

\begin{table*}[!t]
\centering
\begin{tabular}{lccccc}
\toprule
\textbf{Models} & \textbf{Atom} & \textbf{Single-Turn} & \textbf{Multi-Turn} & \textbf{Similar API} & \textbf{Preference}  \\
\midrule
\textbf{GPT-4o-2024-11-20} & \textbf{90.0} & 78.0 & 68.0 & \textbf{80.0} & \textbf{78.0}  \\
\textbf{Llama3.1-70B-Instruct} & 83.7 & 71.5 & 61.0 & 74.0 & 66.0  \\
\textbf{ToolACE-MT-8B} & 83.0 & 64.0 & 51.0 & 68.0 & 68.0 \\
\midrule
\textbf{Qwen2.5-Coder-7B-Instruct}&78.7 &66.5 &51.0 &66.0&40.0\\
\textbf{Llama3.2-3B-Instruct} & 27.0 & 19.0 & 7.0 & 38.0 & 30.0 \\
\midrule
\textbf{BalanceSFT-7B} & 86.0 & \textbf{80.5} & \textbf{74.0} & 70.0 & 56.0  \\
\textbf{BalanceSFT-3B} & 64.3 & 42.0 & 44.0 & 60.0 & 34.0 \\
\bottomrule
\end{tabular}
\caption{Full result of the accuracy of ACEBench normal section. The best result within each category is highlighted in \textbf{bold}.}
\label{ACEfull}
\end{table*}

Table~\ref{ACEfull} presents the comprehensive results of various models on the ACEBench normal section, encompassing multiple evaluation dimensions. For the single-turn task, BalanceSFT-7B stands out with the highest accuracy of 80.5. In the multi-turn task, BalanceSFT-7B again performs best, obtaining an accuracy of 74.0. 




\subsection{Results on Qwen3}

To further verify the effectiveness of BalanceSFT, we conduct experiments by using the more recently open-sourced Qwen3-4B-Instruct-2507. The detailed results are shown in Table~\ref{tab:bfclqwen3}. Compared with the standard SFT method, the proposed BalanceSFT still performs well on the Qwen3 model.

\begin{table*}[!h]
\small
\centering
\begin{tabular}{@{}lccccccccccc@{}}
\toprule
&& \multicolumn{5}{c}{\textbf{Multi-Turn}}& \multicolumn{3}{c}{\textbf{Single-Turn}}\\
\cmidrule(lr){3-7}\cmidrule(lr){8-10}
\textbf{Models}&\begin{tabular}[c]{@{}c@{}}\textit{Parameter}\\ \textit{Counts}\end{tabular}
&\multicolumn{1}{c}{\textit{Overall}} &
  \multicolumn{1}{c}{\textit{Base}} &
  \multicolumn{1}{c}{\begin{tabular}[c]{@{}c@{}}\textit{Miss}\\ \textit{Func}\end{tabular}} &
  \multicolumn{1}{c}{\begin{tabular}[c]{@{}c@{}}\textit{Miss}\\ \textit{Param}\end{tabular}} &
  \multicolumn{1}{c}{\begin{tabular}[c]{@{}c@{}}\textit{Long}\\ \textit{Context}\end{tabular}}&
  \multicolumn{1}{c}{\textit{Overall}} &
  \multicolumn{1}{c}{\textit{Non-Live}} &
  \multicolumn{1}{c}{\textit{Live}} \\ 
  \midrule
\textbf{Qwen3-4B-Inst-2507}&4B &15.75	&19.00	&15.50	&12.50	&16.00&78.19	&86.35	&70.02\\
\textbf{w/ SFT}&4B&41.88&47.00& 50.50& 36.00 &34.00&82.20&85.15&79.25\\
\textbf{w/ BalanceSFT}&4B&47.12 &53.00& 46.00& 41.50&48.00&83.88&87.48&80.28\\
\bottomrule
\end{tabular}
\caption{\label{tab:bfclqwen3} Performance of Qwen3-4B-Instruct-2507 on BFCL.}
\end{table*}

\section{Details of Hard Data Re-sampling Strategy}

In the pipeline of data refinement, we utilize QwQ-32B to assist in different levels of identification and correction. The prompts are shown as follows.

\begin{figure*}[htb] 
\centering
\begin{tcolorbox}[
    colback=gray!5!white,
    title={Response Identification},
    boxrule=0.3mm,
    width=0.95\textwidth,
    arc=3mm,
    auto outer arc=true
]
\textbf{Workflow}

1) Determine whether the Reference Answer is a function\_tool call or a response statement. The function call may not be in the standard function\_tool call format. If it is a function\_tool call, output \texttt{<judge>True</judge>}; if it is a response, output \texttt{<judge>False</judge>}.

\textit{Hint:} Common function\_tool call formats include the following, where \texttt{function\_name} is generally from \texttt{candidate\_function\_tools}:

\texttt{[func\_name1(params\_name1=params\_value1, params\_name2=params\_value2...), \\ func\_name2(params\_name3=params\_value3, params\_name4=params\_value4...)]}

\textbf{Output format:} \texttt{<think>thought process</think><judge>True/False</judge>}

\textbf{Input}

Reference Answer: \texttt{<refANS>}
\end{tcolorbox}
\caption{Response Identification Prompt}
\label{fig:response_identification}
\end{figure*}

\begin{figure*}[htb] 
\centering
\begin{tcolorbox}[
    colback=gray!5!white,
    title={Query and Tool Identification},
    boxrule=0.3mm,
    width=0.95\textwidth,
    arc=3mm,
    auto outer arc=true
]
\textbf{Workflow}

Judge whether the User Query and Candidate Function Tools meet the following requirements:

1) Determine whether the parameter values of the function call can be analyzed from the User Query and the function name can be analyzed from the Candidate Function Tools. If parameter values and function names can be analyzed, output \texttt{<judge>True</judge>}; otherwise, output \texttt{<judge>False</judge>}.

\textbf{Output format:} \texttt{<think>thought process</think><judge>True/False</judge>}

\textbf{Input}

User Query: \texttt{<query>}

Candidate Function Tools: \texttt{<tools>}
\end{tcolorbox}
\caption{Query and Tool Identification Prompt}
\label{fig:query_tool_identification}
\end{figure*}

\begin{figure*}[htb] 
\centering
\begin{tcolorbox}[
    colback=gray!5!white,
    title={CoT Identification},
    boxrule=0.3mm,
    width=0.95\textwidth,
    arc=3mm,
    auto outer arc=true
]
\textbf{Workflow}

Judge whether the Chain-of-Thought leads to the Reference Function Call by evaluating the following requirements:

1) Determine whether the Chain-of-Thought starts from a proper position.

2) Determine whether every step tightly follows the above step and makes correct inference.

3) Determine whether the last step of reasoning points to the correct answer.

If the Chain-of-Thought meets all the requirements, output \texttt{<judge>True</judge>}; otherwise, output \texttt{<judge>False</judge>}.

\textbf{Output format:} \texttt{<think>thought process</think><judge>True/False</judge>}

\textbf{Input}

Chain-of-Thought: \texttt{<CoT process>}

Reference Function Call: \texttt{<refFC>}
\end{tcolorbox}
\caption{Chain-of-Thought Identification Prompt}
\label{fig:cot_identification}
\end{figure*}

\begin{figure*}[htb] 
\centering
\begin{tcolorbox}[
    colback=gray!5!white,
    title={Function and Parameter Identification},
    boxrule=0.3mm,
    width=0.95\textwidth,
    arc=3mm,
    auto outer arc=true
]
\textbf{Workflow}

Judge whether the function names and parameters in the Reference Function Call meet the following requirements:

1) Determine whether the function names and parameter values of the function call are correct. If the function names and parameters are valid, output \texttt{<judge>True</judge>}; otherwise, output \texttt{<judge>False</judge>}.

2) When the function names and parameters requirements are met, output the new function call result according to the Reference Function Call. When the function names and parameters requirements are not met, modify the Reference Function Call format to obtain a new function call result that meets the name and parameters requirements.

\textbf{Output format:}

\texttt{<think>thought process</think><judge>True/False</judge>}

\texttt{<NewFC>}

\texttt{[func\_name1(params\_name1=params\_value1, params\_name2=params\_value2...), \\ func\_name2(params\_name3=params\_value3, params\_name4=params\_value4...)]}

\texttt{</NewFC>}

\textbf{Input}

User query: \texttt{<query>}

Candidate function tools: \texttt{<tools>}

Reference Function Call: \texttt{<refFC>}
\end{tcolorbox}
\caption{Function and Parameter Identification Prompt}
\label{fig:function_parameter_identification}
\end{figure*}

\begin{figure*}[htb] 
\centering
\begin{tcolorbox}[
    colback=gray!5!white,
    title={Format Identification},
    boxrule=0.3mm,
    width=0.95\textwidth,
    arc=3mm,
    auto outer arc=true
]
\textbf{Workflow}

Judge whether the answer meets the following requirements:

1) The output function\_tool format must satisfy the format:

\texttt{[func\_name1(params\_name1=params\_value1, params\_name2=params\_value2...), \\ func\_name2(params\_name3=params\_value3, params\_name4=params\_value4...)]}

\textit{Note:} The parameter name \texttt{param\_name} should not be enclosed in \texttt{"} or \texttt{'}, for example:

\textbf{Incorrect:}
\begin{itemize}
    \item \texttt{[getPrivacyViolationRisk(data="paramvalue1", purpose="paramvalue2")]}
    \item \texttt{[getPrivacyViolationRisk(data='paramvalue1', purpose='paramvalue2')]}
\end{itemize}

The \texttt{paramValue} parameter value, based on its value type, needs to be enclosed in quotes if it is a string.

2) When the output meets the format requirements, output \texttt{<judge>True</judge>}; otherwise, output \texttt{<judge>False</judge>}.

3) When the format requirements are met, output the new function call result according to the Reference Function Call. When the format requirements are not met, modify the Reference Function Call format to obtain a new function call result that meets the format requirements. Do not add or modify parameters and parameter values; only modify the output format.

\textbf{Output format:}

\texttt{<think>thought process</think><judge>True/False</judge>}

\texttt{<NewFC>}

\texttt{[func\_name1(params\_name1=params\_value1, params\_name2=params\_value2...), \\ func\_name2(params\_name3=params\_value3, params\_name4=params\_value4...)]}

\texttt{</NewFC>}

\textbf{Input}

User query: \texttt{<query>}

Candidate function tools: \texttt{<tools>}

Reference Function Call: \texttt{<refFC>}
\end{tcolorbox}
\caption{Format Identification Prompt}
\label{fig:format_identification}
\end{figure*}

\begin{figure*}[htb] 
\centering
\begin{tcolorbox}[
    colback=gray!5!white,
    title={Case Study: Ice Hockey Match Schedule Query},
    boxrule=0.3mm,
    width=0.95\textwidth,
    arc=3mm,
    auto outer arc=true,
    fontupper=\small
]
\textcolor{red}{\textbf{System:}} You are an expert in composing functions. You are given a question and a set of possible functions. Based on the question, you will need to make one or more function/tool calls to achieve the purpose. If none of the functions can be used, point it out. If the given question lacks the parameters required by the function, also point it out.

You should only return the function calls in your response. If you decide to invoke any of the function(s), you MUST put it in the format of \texttt{[func\_name1(params\_name1=params\_value1, params\_name2=params\_value2...), func\_name2(params)]}. You SHOULD NOT include any other text in the response.

\textbf{Here is a list of functions in JSON format that you can invoke:}

\begin{small}
\begin{verbatim}
[{"name": "matchschedules",
  "description": "Fetches the schedules of ice hockey matches for 
    a specified date, including match timings, teams, and other 
    relevant information.",
  "parameters": {
    "day": {"description": "The day of the month for which you 
      want to retrieve the schedules (1-31).", "type": "int", 
      "default": 28},
    "year": {"description": "The year for which you want to 
      retrieve the schedules (e.g., 2022).", "type": "int", 
      "default": 2022},
    "month": {"description": "The month for which you want to 
      retrieve the schedules (1-12).", "type": "int", 
      "default": 11}}}]
\end{verbatim}
\end{small}

\textcolor{orange}{\textbf{User:}} I need the schedules of matches happening on February 28, 2024.

\textcolor{blue}{\textbf{Assistant:}} \texttt{<think>} Okay, let's see. The user is asking for the match schedules on February 28, 2024. The function provided is called \texttt{matchschedules}. The parameters it takes are \texttt{day}, \texttt{month}, and \texttt{year}, all integers. The defaults are day 28, month 11, year 2022. The user's date is February 28, 2024. So February is month 2. The day is 28, which is valid. The year is 2024. Since the function parameters allow specifying these, I need to set each parameter to the user's requested values. The defaults are for November 28, 2022, which isn't what the user wants. So I should override the defaults by setting \texttt{day=28}, \texttt{month=2}, \texttt{year=2024}. That should fetch the correct schedules. The function seems appropriate here. No missing parameters, so the function call is possible. \texttt{</think>}

\texttt{[matchschedules(day=28, month=2, year=2024)]}
\end{tcolorbox}
\caption{Example of Function Call Generation with Chain-of-Thought Reasoning}
\label{fig:case_study}
\end{figure*}

\end{document}